# MuFlex: A Scalable, Physics-based Platform for Multi-Building Flexibility Analysis and Coordination

Ziyan Wu, Ivan Korolija, and Rui Tang*

Institute for Environmental Design and Engineering, The Bartlett, University College London, UK

**Abstract**

With the increasing penetration of renewable generation on the power grid, maintaining system balance requires coordinated demand flexibility from aggregations of buildings. Reinforcement learning (RL) has been widely explored for building controls because of its model-free nature. Open-source simulation testbeds are essential not only for training RL agents but also for fairly benchmarking control strategies. However, most building-sector testbeds target single buildings; multi-building platforms are relatively limited and typically rely on simplified models (e.g., Resistance-Capacitance) or data-driven approaches, which lack the ability to fully capture the physical intricacies and intermediate variables necessary for interpreting control performance. Moreover, these platforms often impose fixed inputs, outputs, and model formats, restricting their applicability as benchmarking tools across diverse control scenarios. To address these gaps, MuFlex, a scalable, open-source platform for multi-building flexibility coordination, was developed. MuFlex enables synchronous information exchange and co-simulation across multiple detailed building models programmed in EnergyPlus and Modelica, and adheres to the latest OpenAI Gym interface, providing a modular, standardized RL implementation. The platform's physics-based capabilities and workflow were demonstrated in a case study coordinating demand flexibility across four office buildings using the Soft Actor–Critic (SAC) algorithm. The results show that under four buildings' coordination, SAC effectively reduced the aggregated peak demand by nearly 12% with maintained indoor comfort to ensure the power demand below the threshold. Additionally, the platform's scalability was investigated through computational benchmarking on building clusters with varying sizes, model types, and simulation programs. The platform is released open-source on GitHub: https://github.com/BuildNexusX/MuFlex.

# 1. Introduction

With the rapid expansion of renewables (e.g., wind, solar), their intermittent and unpredictable natures have posed significant challenges to the real-time stability of power grids [1]. Relying exclusively on large-scale energy storage or backup generators entails substantial investment and may undermine the broader goal of decarbonization [2]. Demand-side management (DSM), which aims to enhance grid stability by actively reshaping end-use electric loads, has become increasingly essential [3]. The building sector accounts for approximately 40% of global energy consumption, making it as an important role at the demand side of power grids [4]. Heating, ventilation, and air-conditioning (HVAC) systems are among the largest energy consumers in buildings, typically accounting for about 30% of total electricity use [5]. In addition to working with active energy systems such as photovoltaics (PV) and batteries, the thermal inertia of buildings can be used as an 'inherent thermal battery' to provide energy flexibility through the methods of precooling, preheating, or setpoint adjustment [6].

In urban distribution networks, critical issues such as peak loads and voltage violations are often concentrated at feeder nodes where buildings are densely interconnected [7]. Aggregating multiple buildings as a single dispatchable entity not only provides sufficient load capacity, but also leverages the temporal diversity of their load profiles, allowing uniform and flexible resource management [8]. As a result, building clusters have become the preferred spatial scale for implementing demand-side management (DSM) strategies recently [9,10]. When HVAC systems of individual buildings in a cluster independently respond to the demand response (DR) signal from power grids, the aggregated effects cannot be ensured as expected without the proper coordination across buildings [10]. At the level of multi-buildings, an effective coordination mechanism is therefore essential to secure the aggregated effects of a building cluster, while accounting for each building's objective [11].

Currently, rule-based feedback control is dominant in the control of HVAC systems. Its control logic normally adheres to a pre-determined schedule set by experienced engineers or guidelines (e.g., ASHRAE Guideline 36) [12]. But control actions based on fixed rules often lack responsiveness to changing operational conditions and requests from power grids, posing the risk of significant deviations from optimal building operations. Advanced control techniques can overcome the limitations of rule-based controls and are generally categorized into two methods: model predictive control (MPC) and reinforcement learning (RL) [13]. MPC uses receding-horizon optimization with a model to capture building dynamics and decide optimal control actions, but the modelling process is time-consuming and requires significant expertise, and the control performance is highly sensitive to the model accuracy [14]. RL, on the other hand, does not require an explicit dynamics model (e.g., transition functions) to decide optimal control actions. Instead, it learns a policy through trial-and-error interaction with an environment, which can be either a model or a real building. Due to this attractive "model-free" nature, RL has demonstrated strong potential across a wide range of decision-making tasks [15]. Advances in RL in other fields have been driven by the availability of rich, standardized environments that enable direct comparison and evaluation of newly developed algorithms, such as games [16], robotics [17] and traffic management [18]. In the building sector, researchers are increasingly turning to open-source simulation testbeds to train and benchmark RL controllers, as a practical path to demonstrate their benefits under realistic operating conditions. A common workflow is to pre-train RL policies offline in simulation, evaluate them under varied scenarios, and then deploy them online in real buildings [13]. This can support the rapid development of algorithms, fair cross-comparison, and real-world deployment [19]. However, most of these efforts are focused on the single-building level. At the multi-building level, the platform applicable to benchmark control strategies, providing comprehensive analyses of control performance across building clusters, is limited. The absence of a standardized platform leads to developed algorithms being tested only under customized scenarios and baselines, which makes it difficult to accurately quantify performance.

In this paper, therefore, an open-source physics-based testbed, MuFlex, was developed for training and benchmarking control strategies for multi-building flexibility coordination. The contributions of this work include:

(1) The building systems within the platform are modeled using physics-based white-box models (i.e., EnergyPlus and Modelica models), which provide a detailed physical representation of building dynamics. It allows flexible exposure of control variables (i.e., model inputs) and system states (i.e., model outputs) as needed. The inputs can be defined as real actuator variables in the HVAC system, enabling zone-level controls; the outputs can include intermediate system variables, which supports clear interpretation of control strategies.

(2) A communication hub is developed to enable real-time interaction between multiple building models and control agents using the Functional Mock-up Interface (FMI) protocol. It can synchronize the co-simulation of multiple models from different simulation programs, and leverages parallel execution of models for computational efficiency.

(3) The control agent is embedded in a standardized RL environment based on the latest OpenAI Gymnasium API [20], thereby enabling use with Stable-Baselines3 [21] and recent standard RL libraries.

(4) The platform's physics-based capabilities and workflow were demonstrated through the implementation of the Soft Actor-Critic (SAC) algorithm with comprehensive hyperparameter tuning and effective strategies of action space discrete mapping for the coordination of four buildings of energy use. Detailed zone-level control was applied for each building to interpret the resulting controls.

(5) The platform's scalability was investigated through computational benchmarking on building clusters of 4, 20, and 50 buildings in terms of runtime and memory usage. The buildings included different types and were modeled by EnergyPlus and Modelica.

## 2. Literature Review

Reinforcement learning (RL) is a subset of machine learning that learns to obtain optimal sequences of actions through interacting with a dynamic environment. By adjusting its policy to output the actions that can maximize the cumulative rewards, RL provides remarkable model-free characteristics and is promising to benefit building controls [22]. Many studies have been conducted on applying RL for single building optimization [23,24]. In the development and test of RL algorithms, simulation testbeds are essential. The white-box models, such as EnergyPlus [25] and Modelica models [26], can provide detailed building dynamics and description, which are normally used to interact with control agents/algorithms [27]. Typically, there are four ways to enable communication between the RL algorithm (control agent) and physical building models, including EnergyPlus Energy Management System (EMS) [28], EnergyPlus Python Application Programming Interface (API) [29], Building Controls Virtual Test Bed (BCVTB) [30] and Functional Mock-up Interface (FMI) [31]. EnergyPlus EMS is an EnergyPlus built-in control script, while EnergyPlus Python API is an external extension interface designed for EnergyPlus and Python interaction. They are regarded as a one-to-one approach tied exclusively to EnergyPlus. BCVTB acts as a middleware connecting different models and programs, and is capable of supporting multiple programs' interconnection. FMI offers direct coupling between different programs with a standardized interface.

Based on these mechanisms, the open-source platforms for single-building RL development are summarized in Table 1. COBS [32] extracts EnergyPlus sensor data through EMS and implements an event queue that follows the standard RL interaction paradigm. The rl-testbed for energyplus [33] offers a data-center EnergyPlus model and enables interaction by adding two built-in functions to the EMS scripts. The rllib-energyplus [34] and Sinergym [35] both communicate with EnergyPlus via the wrapped EnergyPlus Python API; Sinergym is actively maintained with the latest Gymnasium API. Gym_Eplus [36] uses the ExternalInterface function of EnergyPlus to connect control scripts through BCVTB, but it relies on an older EnergyPlus version without active maintenance. As Modelica relies on equation-based modelling and can generate analytic derivatives, it can provide gradient-based optimization. Most platforms using Modelica models adopt FMI, which can also support the interactions with EnergyPlus. Energym [37] includes 14 models in Modelica and EnergyPlus, some of which incorporate models of batteries and electric vehicles. AlphaBuilding Medoffice [38] targets zone-level controls of a Variable Air Volume (VAV) system in a medium office building. FlexDRL [39] is built on a real experimental building and constructs a co-simulation environment combining an EnergyPlus envelope model with Modelica PV and battery models. Advanced Controls Test Bed [40] employs Spawn of EnergyPlus, providing a model-exchange framework with building envelope and internal gains in EnergyPlus while HVAC systems and controls in Modelica. BOPTEST-gym [41] implements a RESTful HTTP API based on FMI that runs diverse building test cases on servers and also integrates with latest Gymnasium.

Table 1 Open-source simulation testbeds for a single building

| **Name** | **Building Model** | **Communication** | **Latest Update** |
|---|---|---|---|
| COBS [32] | EnergyPlus | EnergyPlus EMS | 2022 |
| rl-testbed for energyplus [33] | EnergyPlus | EnergyPlus EMS | 2023 |
| rllib-energyplus [34] | EnergyPlus | EnergyPlus Python API | 2024 |
| Sinergym [35] | EnergyPlus | EnergyPlus Python API | 2025 |
| Gym_Eplus [36] | EnergyPlus | BCVTB | 2019 |
| Energym [37] | EnergyPlus/Modelica | FMI | 2021 |
| AlphaBuilding Medoffice [38] | EnergyPlus/Modelica | FMI | 2021 |
| FlexDRL [39] | EnergyPlus/Modelica | FMI | 2022 |
| Advanced Controls Test Bed [40] | EnergyPlus/Modelica | FMI | 2022 |
| BOPTEST-gym [41] | Modelica | FMI | 2025 |

Many benchmarking studies at the single-building level are conducted using these open-source platforms to identify the outperformed RL algorithms as candidates for real-world deployment. Gao and Wang [42] used the BOPTEST-gym to evaluate Soft Actor-Critic (SAC), Double Deep Q-Network (DDQN) and a Dyna-style model-based RL (MBRL) approach and showed that MBRL can match the performance of model-free methods with substantially reduced training time. Wu et al. [43] performed a comprehensive benchmark of five different categories of RL algorithms using Gym_Eplus, analyzing their learning progress, convergence and stability. Wang et al. [44] benchmarked MPC and three RL algorithms on BOPTEST-gym on control performance, data efficiency, implementation efforts and computational demands. These benchmarking results are informative for real-world deployment, as single-building testbeds are typically built on physics-based models that provide rich intermediate states for interpreting control performance and improving policy transfer to real buildings.

As buildings transition from energy consumers into assets for demand-side management, control algorithms are not only required to realize individual building-level optimization, but also coordinately provide flexibility to the grid across multiple buildings [22]. Tang et al. [45] pointed out that the occurrence times of peak demand in individual buildings were always different from that of aggregated peak demand. Uncoordinated control therefore induced potential new peaks and wasted efforts. Shen et al. [46] observed that price-based demand response (DR) programs could cause new undesirable peak demands during periods with low electricity prices if without properly coordinated across buildings. Since the scale of buildings increases the complexity of the control problem [47], some of studies fix building energy demand to simplify the control problems at multi-building level and only focus on optimizing the power dispatching of energy systems (e.g., PV, storage) to track a pre-determined (fixed) building load profile, which is normally derived from a fixed schedule of building system operation such as indoor temperature setpoints [48,49]. CityLearn v1 [50] was developed to benchmark the control performance of multi-building systems cooperating with distributed energy systems, which used hourly energy demand of buildings from pre-simulated EnergyPlus models. Pigott et al. [51] proposed GridLearn, an adaptation of CityLearn for the evaluation of voltage levels across the distribution grid, with the assumption of constant thermal demands. Another simplification treats residential demand as a modifiable appliance load, typically considering only the need to meet the demand within a specified time window, while ignoring the resulting changes in indoor conditions [52,53]. These studies attempt to simplify the control process, making the work different from real conditions and incapable of capturing building dynamics to leverage building thermal mass and indoor temperature reset as flexibility sources.

To incorporate building dynamics at the multi-building level in the demand side management, studies rely on two main approaches: data-driven models (i.e., black-box modelling) and resistance and capacitance (RC) models (i.e., grey-box modelling). Zhang et al. [54] adopted an autoregressive model for four small offices to infer the indoor temperature of each zone. Pinto et al. [55] extended CityLearn by integrating a Long Short-Term Memory (LSTM) network to model building thermal dynamics and using the platform to control the heat pumps of a four-building cluster, facilitating the development of CityLearn v2. Fonseca et al. [56] further extended the CityLearn v2 by modelling Electric Vehicles (EVs), their charging infrastructure and associated energy flexibility dynamics. Gallo & Capozzoli [57] simulated a residential community of 50 households, each equipped with heat pump, PV, and thermal energy storage. Wang et al. [58] developed AlphaBuilding ResCommunity, providing on/off control of a heat pump in each household and demonstrating scalability in a case study of 1000 households.

Even though black-box or grey-box models to simulate building thermal dynamics offer computational convenience, they introduce limitations on existing multi-building simulation:

(1) The model interfaces of existing multi-building platforms are often inflexible with fixed inputs and outputs. For example, if different variables are expected to control or observe, RC models typically require recalibration and parameter identification, while data-driven models require extra retraining.

(2) RC and data-driven models do not explicitly represent the building dynamics and HVAC operations and therefore cannot provide intermediate variables needed to fully interpret the resulting control performance.

(3) Existing multi-building simulation commonly represent each building as a single thermal zone and provide simplified building-level control, which cannot support zone-level observation to provide sufficient control details of individual buildings.

Therefore, the fixed and limited inputs, outputs, and model formats of these platforms restrict their applicability as benchmarking tools for diverse control scenarios to support the research and development of advanced control strategies across multiple buildings. In addition, in contrast to single-building control systems, there are very few platforms or environments available that offer a comparable level of accuracy and applicability for multi-building contexts. It is essential to develop a multi-building simulation platform that can effectively address the above limitations.

# 3. Development of Multi-building Control Platform, MuFlex

## 3.1 Framework of the platform

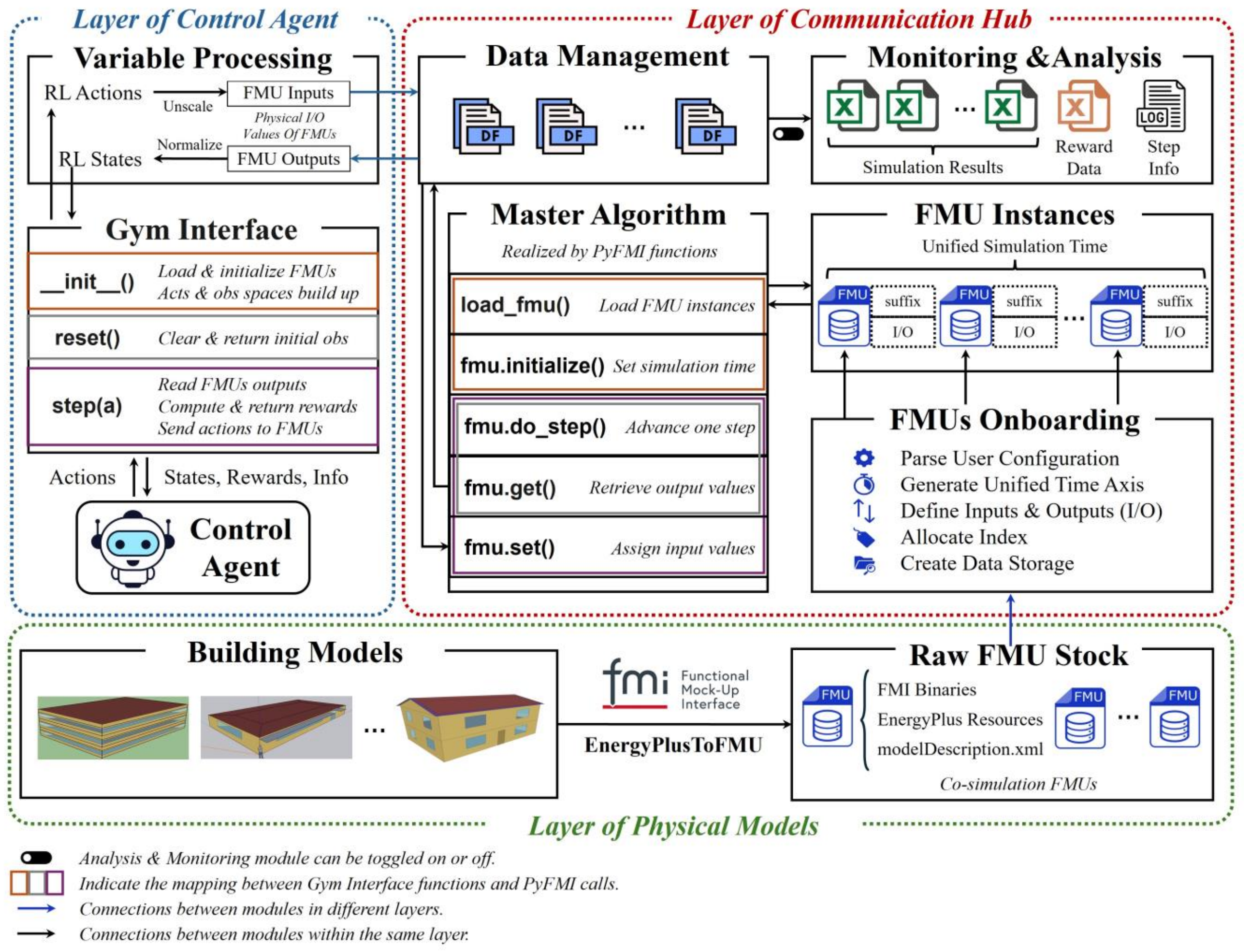

Fig. 1. The platform structure, MuFlex

The overall structure of platform MuFlex is shown in Figure 1. It consists of three parts: the physical models, the communication hub, and the control agent. In the layer of physical models, building stock models (i.e., EnergyPlus models) are converted into raw FMUs. In the layer of control agent, the standard Gym interface facilitates interaction with the control agent, translating the agent's actions into physical control signals and receiving model outputs as observations. The communication hub manipulates the FMUs and enables real-time interaction between the control agent and building models, while recording the data.

### 3.2 The layer of Physical Models

The layer of physical models provides the white-box models of buildings. As discussed in the literature review, among three primary approaches to enable the interaction between physical models and control agents, the approach of employing the Functional Mock-up Interface (FMI) is selected to manipulate the models in the platform because of its efficiency and scalability. FMI has been adopted by more than 40 modelling and simulation environments and is under active development and maintenance [59].

In the platform, the FMI standard packages a simulator together with its solver and metadata into a zipped archive with the .fmu extension, i.e., Functional Mock-up Units (FMUs). Co-Simulation FMUs encapsulate their own solvers, and the master algorithm coordinates data exchange and synchronizes time between self-contained FMUs. In the layer of physical models, each building is modelled by EnergyPlus and exposed through the ExternalInterface to convert to an FMU with EnergyPlusToFMU. For Modelica building models, each model is exported as an FMU using an FMI-compatible Modelica software (e.g., Dymola or OpenModelica). Taking EnergyPlus models as an example, each converted raw FMU is a self-contained archive that bundles three essential elements including FMI binaries, the EnergyPlus resources directory and a modelDescription.xml file. FMI binaries are pre-compiled dynamic-link libraries (i.e., DLLs on Windows, .so files on Linux) that encapsulate the core solver required for running EnergyPlus through the FMI interface. The EnergyPlus resources directory is a bundle of all runtime assets and configurations, including the primary IDF input file, the EPW weather file and any auxiliary data the model may reference. The modelDescription.xml file, defined by the FMI specification, enumerates every exchangeable variable together with metadata such as causality, units, default timestep size, and version information. The raw FMUs are subsequently loaded by the Communication Hub into the Python environment. Using FMI keeps the developed platform open to integrate other programs and extend its functionalities.

Additionally, FMI streamlines the simulation setup as each model is delivered as an archive. Simulations can be run simply by placing the raw FMU files in a single directory recognized by the hub with no need for auxiliary scripts or configuration. Accordingly, integrating a new model becomes a plug-and-play process. If a building model is already packaged as an FMU (e.g., from other FMI-based single-building testbeds, such as Energym [37]), it can be incorporated by (i) placing the FMU file in the platform-recognized model directory and (ii) providing the variable names for model inputs and outputs. For raw models that are not yet FMU-packaged, they can first be exported to FMUs following the workflow described above and then integrated in the same manner. The detailed Inputs/Outputs (I/O) definitions are introduced in the following section.

### 3.3 The layer of Communication Hub

The Communication Hub is responsible for processing the raw FMUs, running them synchronously, processing received inputs from the control agent, sending control signals back for the next timestep, and also recording and printing the data at each timestep. The open-source Python library PyFMI [60] is deployed in the platform to enable the interactions across FMUs of individual buildings. The master algorithm employs five FMU functions in PyFMI, including loading the FMU instances with *load_fmu()*; setting the simulation time with *fmu.initialize()*; performing a simulation step with *fmu.do_step()*; retrieving and recording outputs after each step with *fmu.get()*; and assigning new inputs from the control agent for the next step with *fmu.set()*. These functions need to be correctly called using the specified FMU instance together with the variable names and values defined in the .xml file. To make a multi-FMU (multi-building) scenario operational and easily organized, five processing steps are implemented for the raw FMUs from the layer of physical models before using PyFMI functions to operate the FMUs and set up the interaction with the control agent:

**(1) Parse User Configuration**: This involves reading the FMU configurations (e.g., path, type, number), simulation settings (e.g., duration, timestep), scenarios (e.g., action space type, reward function parameters) and toggle switches for automatic data storage and visualization defined by the user. This centralized parsing of user inputs minimizes redundant checks and provides a user-friendly interface for customizing control scenarios, particularly enabling flexible configuration of building clusters with varying numbers and types of buildings.

**(2) Generate Unified Time Axis**: A global time grid is constructed for all FMUs to ensure synchronized multi-building simulation. It is used by the PyFMI functions *fmu.initialize()* and *fmu.do_step()* to drive the co-simulation. This time axis also serves as the unified clock across multiple modules in the platform including generating 'hour of day' information of observations and indexing data storage, preventing time misalignment between FMUs and other modules.

**(3) Define Inputs & Outputs (I/O)**: As multiple FMUs introduce a large number of variables, a modular I/O system is developed to efficiently manage data, including inputs, outputs, discretization granularity, and physical upper/lower bounds. Each group of FMUs with the same I/O is mapped to one dictionary entry in the I/O system.

When calling the PyFMI functions *fmu.get()* and *fmu.set()*, the appropriate variable name for each FMU can be retrieved automatically from that system. The discretization granularity and the physical upper and lower bounds are used to construct the discrete or continuous action spaces in Gymnasium and define the scaling for normalization of actions and observations for the control agent. This design enables straightforward scaling of the cluster by reusing shared I/O definitions for identical building archetypes, while also allowing new models with different I/O definitions to be added by registering an additional I/O entry with minimal configuration effort.

**(4) Allocate Index**: PyFMI functions require specifying the corresponding FMU instance to ensure the independence of parallel operations. For this reason, the platform uses an index *fmu_index* (e.g., *fmu_1*, *fmu_2*, *fmu_3,* etc.) to uniquely identify and isolate each building model in the order defined by the user's FMU configuration. The index also differentiates the same variable names across FMUs.

**(5) Create Data Storage**: For each FMU, a DataFrame is created to record and store simulation data (inputs and outputs) at each timestep, centralizing the management of physical values passed to *fmu.set()* and retrieved via *fmu.get()*. A separate reward list is used to record the breakdown of reward components per step. This enables post-simulation data export for result visualization, debugging, and analysis of control strategies. During runtime, the platform can print log messages to the terminal, reporting FMU status, step-by-step simulation data, and any warnings or errors using the data management module.

Through these five steps, the FMUs are fully prepared for execution by the master algorithm. At each timestep, following the FMU index, the first FMU has its type identified and the corresponding I/O configuration retrieved to obtain the list of output variable names. This list is passed to *fmu.get()* to retrieve all outputs for the current timestep, and these values are recorded in a DataFrame. The same procedure is repeated for the other FMUs. Once the outputs of all FMUs have been recorded, the data stored in DataFrames are concatenated into an observation vector, which is sent to the control agent layer for normalization. The control agent returns a list whose length matches the number of FMUs, with each entry containing a sub-list of physical setpoints corresponding to a specific FMU. Based on the FMU index, these action segments are matched to each FMU along with their respective input variable name lists. The *fmu.set()* function is then called to pass the new setpoints into the FMUs.

### 3.4 The layer of Control Agent

The control agent layer in the platform is designed to support not only the implementation of diverse RL algorithms, but also the deployment of rule-based control and Model Predictive Control (MPC). This versatility maximizes the platform's applicability and establishes it as a standardized benchmarking environment for a wide range of control methodologies. The Gymnasium API [20] (formerly OpenAI Gym [61]) as the latest Gymnasium interface, is applied to enable the developed platform to integrate with recent RL algorithm libraries such as Stable-Baselines3 [21] or RLlib [62]. The interface specifies three core functions, *__init__()*, *reset()* and *step(a)*, along with data structures that describe the action and state spaces, to interact seamlessly with an environment.

Note that the Communication Hub works with the FMU's actual physical values while the RL agent operates on normalized ones to improve training stability. After concatenating all FMU outputs into the observation vector, a variable processing module scales it into the [0, 1] range before feeding it to the agent as states, which then returns discrete or continuous actions depending on the chosen algorithm. The environment provides two action spaces: MultiDiscrete (a one-dimensional integer array, each element indicating the bin index for a control input) and Box (a one-dimensional float array in $[-1, 1]$). The length of the arrays for both action types equals the total number of control inputs across all FMUs. The processing module then unscales these actions into real physical setpoints and passes them to the Communication Hub. The unscaling mechanism will be further discussed in the next section 4.2.

The control agent is insulated from the underlying PyFMI API and interacts exclusively with the Gym interface. In Figure 1, the implementation of each interface function in the control agent is visualized with distinct colored boxes, linking PyFMI functions with Gymnasium methods to form a clear call chain. Specifically, the *_init_()* method includes loading and initialization of FMUs, as well as the construction of the RL action and observation spaces. When *reset()* is called, it clears all logs and internal counters from the previous run, rolls back the simulation time to the initial moment, and returns the first normalized observation. The execution of *step(a)* provides the values of control actions and initiates the next step.

# 4. Platform Demonstration and Case Study

## 4.1 The settings of case study

To demonstrate the functionalities of the platform, four buildings, including two small and two medium office buildings, are formulated as a group to coordinate their power demand for the service of power grids. The details of these four buildings are shown in Table 2. Small Office A and B share the same layout with a total floor area of 5,500 $ft^2$ and Medium Office A and B share the same layout with a floor area of 53,628 $ft^2$. Small Office B and Medium Office B maintain a constant occupancy fraction throughout the day, while Small Office A and Medium Office A exhibit varying degrees of occupancy reduction around lunchtime. As shown in Table 2, the zone and floor details comprise a top view illustrating the distribution of five zones in the small office layout, and a 3D schematic of the three floors in the medium office layout, with the number of occupants in each zone or floor indicated in parentheses. The spatial and temporal differences, together with variations in heating, ventilation, and air conditioning (HVAC) system configuration, give rise to distinct power demand profiles of HVAC systems across four buildings. The HVAC systems for buildings are the packaged Variable Air Volume (VAV) system served by an air handling unit (AHU) equipped with a direct-expansion (DX) cooling coil. For cooling season, the DX coil cools the supply air by refrigerant evaporating directly within the coil. The schematic of the VAV system used in small offices is presented in Figure 2. Each zone is served by a VAV terminal box with a modulating damper that adjusts the zone supply airflow. An outdoor air damper mixes outside fresh air with return air from the zones to meet ventilation requirements. In small offices, a single VAV system serves five zones; in medium offices, three identical VAV systems independently serve each floor. These models were built in EnergyPlus V9.2.0, and EnergyPlusToFMU V3.1.0 [63] was used accordingly to export them as co-simulation FMUs in the platform. A hot summer day (20th July) in Nanjing, China was selected as the test day with ambient temperatures ranging from 28.6 °C (at 6 AM) to 37.2 °C (at 4 PM). Twelve days preceding July 20th, i.e., July 8th -19th, were used as training episode, since the daily maximum temperature first exceeded 30 degrees on July 8th and reached its annual peak on July 20th, as shown in Figure 3.

The control timestep for updating the optimized setpoints was set as 15 minutes, and the acceptable range of indoor temperatures of buildings was 23-25 °C. For the air handling unit (AHU) supply air temperature, the valid range was set as 10-15 °C. The temperature precision was 0.1 °C. The simulation was performed on a laptop, equipped with a 12th-Gen Intel Core i5-12400F CPU and 32 GB DDR4 memory (3200 MHz). The GPU was NVIDIA GeForce RTX 4060 Ti with 8 GB VRAM. GPU acceleration has minimal impact on simulation runtime (i.e., EnergyPlus, Modelica), but it can substantially affect RL training time.

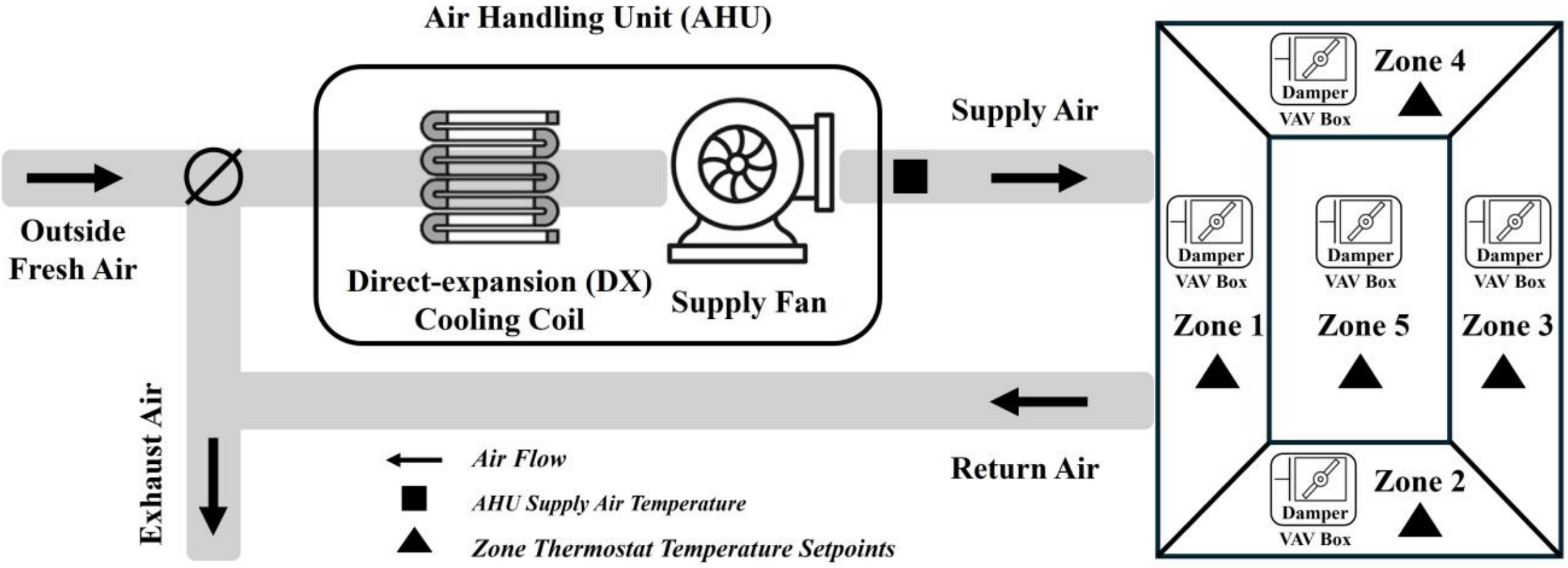

Fig. 2. Schematic of the VAV system used in small offices

Table 2 Four office buildings for the test and demonstration

| | **Small Office A** | **Small Office B** | **Medium Office A** | **Medium Office B** |
|---|---|---|---|---|
| Building Geometry | | | | |
| Floor Area (ft$^2$) | 5,500 | 5,500 | 53,628 | 53,628 |
| Occupancy Fraction | | | | |
| Zone/Floor Layout (* the number in parenthesis is the maximum occupancy number) | Zone3 (10)<br>Zone5 (21)<br>Zone1 (10)<br>Zone4 (6)<br>Zone2 (6) | Zone3 (11)<br>Zone5 (20)<br>Zone1 (11)<br>Zone4 (5)<br>Zone2 (5) | TopFloor (31)<br>MidFloor (31)<br>BotFloor (31) | TopFloor (26)<br>MidFloor (26)<br>BotFloor (26) |

Each building has six control variables (i.e., inputs of the FMUs) to manage the power demand. In small offices, the supply air temperature of VAV system and five indoor temperature setpoints of five corresponding zones are controlled. In medium offices, each floor is regarded as a ‘control zone’ and the supply air temperature of VAV system and the temperature setpoint for each of the three floors are controlled. The zone temperature setpoint here refers to the thermostat setpoint. Once a setpoint is specified, the HVAC system tracks this target subject to its system capacity, regulating the zone temperature toward the setpoint. The locations of the selected control variables within the VAV system are annotated in Figure 2. Each zone’s supply airflow is set by the VAV damper based on supply air temperature and the zone setpoint, subject to hard constraints of occupancy-based minimum ventilation. The information provided to the control agent for optimization (i.e., FMU outputs) is outdoor weather conditions, HVAC power and operational variables, and indoor temperatures at each zone (small office) or floor (medium office), as summarized and indexed in Table 3. Small offices have 6 control inputs and 13 outputs, while medium offices have 6 inputs and 19 outputs.

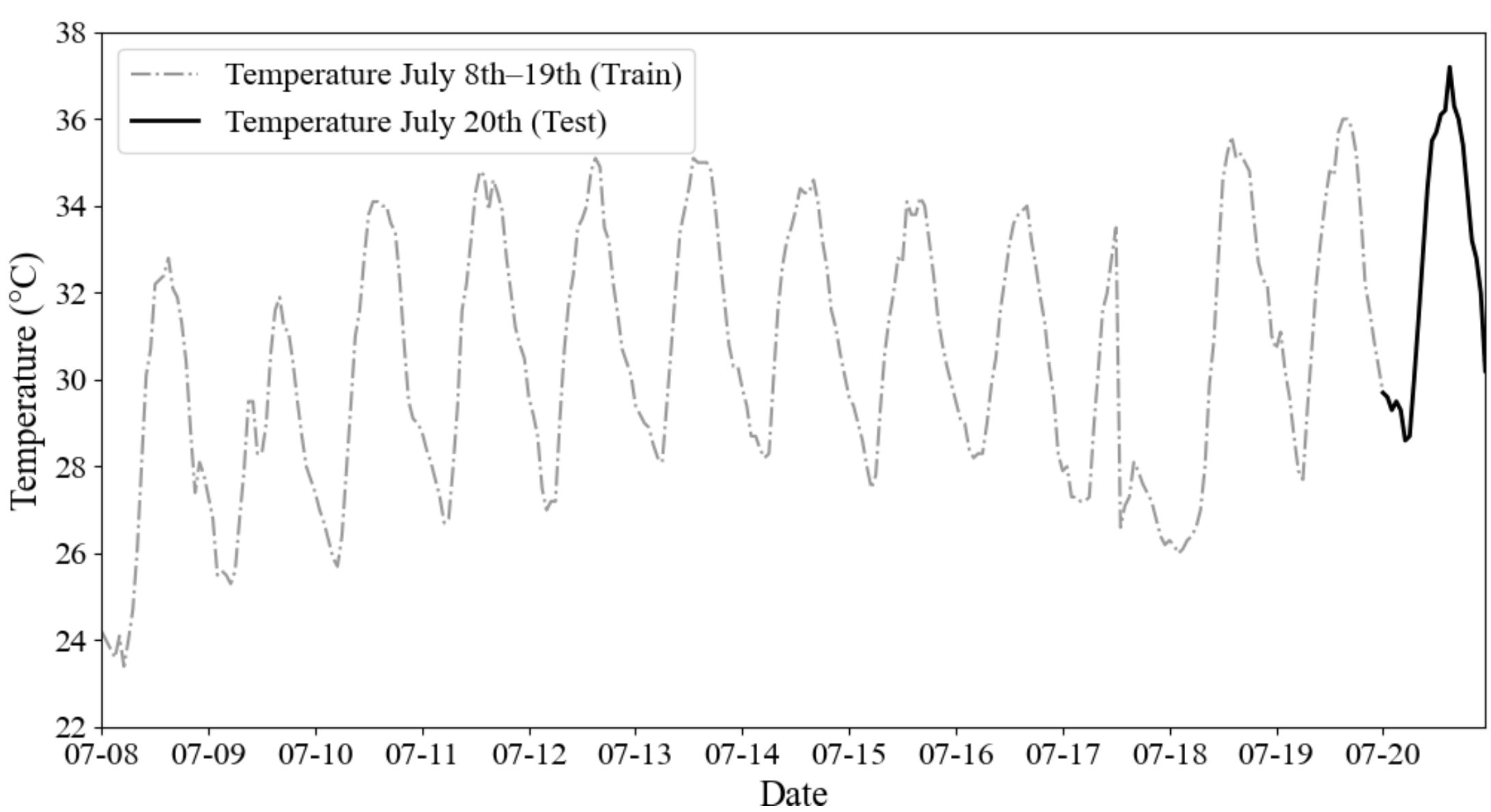

Fig. 3. Outdoor dry-bulb temperature of training and testing periods

Table 3 Inputs and outputs of the FMUs

| | **Small Office I/O No. (index)** | **Variables** | **Medium Office I/O No. (index)** |
|---|---|---|---|
| Inputs | 1 | AHU Supply Air Temperature, °C | 1-3 |
| | 2-6 | Zone/Floor Cooling Setpoint, °C | 4-6 |
| Outputs | 1 | Site Outdoor Air Dry-bulb Temperature, °C | 1 |
| | 2 | Site Outdoor Air Relative Humidity, % | 2 |
| | 3 | Site Wind Speed, m/s | 3 |
| | 4 | Site Direct Solar Radiation Rate per Area, W/m$^2$ | 4 |
| | 5 | Cooling Coil Electric Power, W | 5-7 |
| | 6 | Fan Electric Power, W | 8-10 |
| | 7 | Supply Air temperature, °C | 11-13 |
| | 8 | Supply Air Mass Flow Rate, kg/s | 14-16 |
| | 9-13 | Zone/Floor Indoor Air Temperature, °C | 17-19 |

A rule-based controller (RBC) using a fixed indoor temperature setpoint at 25 °C and supply air temperature of VAV system at 15 °C was set as the baseline controller. The 25 °C setpoint corresponds to the upper bound of the 23–25 °C comfort band and yields a baseline with minimum cooling energy and peak demand, providing a challenging reference for evaluating the RL controller.

### 4.2 The settings of RL control algorithm - Soft Actor-Critic

Coordination among multiple buildings needs to handle information from involved buildings and therefore, make the control problem high-dimensional. Based on the designed inputs and outputs above, the control problem in this study is a 24-dimensional action space (six control variables per building) and a 53-dimensional state space (consisting of 26 small offices outputs, 38 medium offices outputs, elimination of 12 duplicate outdoor weather variables, and the current hour/time information). Model-free Reinforcement Learning (RL) algorithms are sample-inefficient, and high demand for interactions to learn complex behaviors with high-dimensional observations. Soft Actor-Critic (SAC), as an off-policy method, can reuse experience to improve sample efficiency. SAC adds an entropy term to the standard objective of maximizing cumulative reward, so that while seeking high returns, the policy maintains sufficient randomness and diversity to offer two advantages: more thorough exploration in high-dimensional spaces and more robust and less prone to converging prematurely to a rigid suboptimal policy when faced with errors [64].

Specifically, SAC augments the return with an entropy term scaled by a temperature parameter $\alpha$. The soft state-value function is shown in Eq. (1), where the temperature parameter $\alpha$ determines the relative weight of the entropy term versus the reward. A larger $\alpha$ encourages higher-entropy (more stochastic) policies, while a smaller $\alpha$ drives the policy toward pure return maximization. Consistently, the corresponding 'soft' Bellman backup for Q-value is shown in Eq. (2).

$$V_{\pi}(s) = \mathbb{E}_{a \sim \pi}[Q_{\pi}(s,a) - \alpha log\pi(a|s)] \tag{1}$$

$$Q_{\pi}(s,a) = r(s,a) + \gamma \mathbb{E}_{s' \sim p}[V\pi(s')] \tag{2}$$

To reduce over-estimation bias in value backups when using a single Q network [65], double Q networks are employed to improve training stability and convergence in this study. Two slowly updated target critics $\bar{Q}_{\theta_1}$, $\bar{Q}_{\theta_2}$ are maintained, and the minimum of the target Q-values is used inside the soft target. Given a transition $(s,a,r,s')$ sampled from the replay buffer $\mathcal{D}$, each critic $Q_i$ minimizes the loss in Eq. (3). In addition, target critics are updated softly via Polyak averaging in Eq. (4) with a mixing factor $\tau = 5 \times 10^{-3}$ that gently incorporates the latest critic parameters and stabilizes learning.

$$\mathcal{L}_{Q_i} = \mathbb{E}_{(s,a,r,s')\sim D}\left[\left(Q_{\theta_i}(s,a) - \underbrace{\left[r+\gamma\left(min_j \bar{Q}_{\theta_i}(s',a') - \alpha log\pi_{\phi}(a'|s')\right)\right]}_{soft\ target}\right)^2\right], a'\sim\pi_{\phi}(\cdot\,|s') \quad (3)$$

$$\bar{\theta}_i \leftarrow \tau\theta_i + (1-\tau)\bar{\theta}_i \quad (4)$$

SAC updates the policy according to the soft optimality principle in Eq. (5), which favors high soft Q actions while preserving entropy. In practice, actions are sampled via a squashed Gaussian reparameterization in Eq. (6), which lets gradients flow through $\mu_{\phi}$ and $\sigma_{\phi}$. With the required Jacobian correction for the tanh squashing in Eq. (7), this construction yields the actor loss, as shown in Eq. (8).

$$\pi^*(a|s) \propto exp\left(\frac{1}{\alpha}Q^*(s,a)\right) \quad (5)$$

$$a = g_{\phi}(\varepsilon,s) = \tanh\left(\mu_{\phi} + \sigma_{\phi}(s)\varepsilon\right), \varepsilon \sim \mathcal{N}(0,1) \quad (6)$$

$$log\pi_{\phi}(a|s) = log\mathcal{N}\left(\mu_{\phi},\sigma_{\phi}\right) - \sum_i \log\left(1-a_i^2\right) \quad (7)$$

$$\mathcal{L}_{\pi}(\phi) = \mathbb{E}_{s,\varepsilon}\left[\alpha log\pi_{\phi}\left(g_{\phi}(\varepsilon,s)|s\right) - min_j Q_{\theta_j}\left(s,g_{\phi}(\varepsilon,s)\right)\right] \quad (8)$$

According to Haarnoja et al. [66], the SAC may be vulnerable to this temperature hyperparameter. To optimize its value, this study adopted a gradient-based automatic adaptation method for $\alpha$, which could tune the expected entropy of the states to match a target value by parameterizing $\alpha = exp(log\alpha)$ and minimizing the Eq. (9) to drive the empirical policy entropy toward the target $\bar{\mathcal{H}}$, setting to the negative action dimension $-|\mathcal{A}|$.

$$\mathcal{L}_{\alpha}(\alpha) = -(log\alpha)[\mathbb{E}_{a\sim\pi_{\phi}} log\pi_{\phi}(a|s) + \bar{\mathcal{H}}], \bar{\mathcal{H}} = -|\mathcal{A}|(=-24), \quad (9)$$

The effects of double Q networks and automatic temperature tuning will be discussed in Section 5.1. Table 4 shows the workflow of SAC by the pseudocode.

Table 4 Pseudocode of SAC implementation

| | |
|---|---|
| **Input:** $\theta_1, \theta_2, \phi, log\alpha$ | Initial parameters |
| **Init:** $\overline{\theta_1} \leftarrow \theta_1, \overline{\theta_2} \leftarrow \theta_2$ | Copy critic weights to create target networks |
| **Init:** $D \leftarrow \emptyset$ | Create an empty replay buffer |
| **for** each episode **do** | |
| $env.reset()$ | Reset the *MuFlex* environment |
| **while not done do** | |
| $a_t \sim \pi_\phi(a \mid s_t)$ | Sample action from Gaussian-tanh policy |
| $(s_{t+1}, r_t, done) \leftarrow env.step(a_t)$ | Execute action and observe transition |
| $D \leftarrow D \cup \{(s_t, a_t, r_t, s_{t+1}, done)\}$ | Store transition in the replay buffer |
| **if** $\lvert D \rvert \geq batch$ **then** | Start gradient updates |
| $Sample\ B \subset D$ | Draw a mini-batch from the buffer |
| // **Critic target** (Eq. (2)) | |
| $a', log\pi' \sim \pi_\phi(s')$ | Sample next action with current policy |
| $y \leftarrow r + \gamma(1-d)(min(\bar{Q}_1, \bar{Q}_2)(s', a') - \alpha log\pi')$ | Bootstrapped soft value target |
| // **Critic update** (Eq. (3)) | |
| $\theta_i \leftarrow \theta i - \lambda_Q \nabla_{\theta_i} \mathcal{L}_{Q_i}, i \in \{1,2\}$ | Update two Q-networks |
| // **Actor update** (Eq. (8)) | |
| $\hat{a}, log\pi \sim \pi_\phi(s)$ | Resample current action |
| $\phi \leftarrow \phi - \lambda_\pi \nabla_\phi \mathcal{L}_{\pi_\phi}$ | Update actor parameters |
| // **Temperature update** (Eq. (9)) | |
| $log\alpha \leftarrow log\alpha - \lambda_\alpha \nabla_{log\alpha} \mathcal{L}_\alpha$ | Adjust the temperature toward the target entropy |
| // **Soft update** (Eq. (4)) | |
| $\bar{\theta}_i \leftarrow \tau\theta i + (1-\tau)\bar{\theta}_i, i \in \{1,2\}$ | Soft-update target critics |
| **end if** | |
| **end while** | |
| **end for** | |
| **Output:** $\theta_1 *, \theta_2 *, \phi *, log\alpha *$ | Optimized parameters |

## Action and State Space

In this study, the action space and state space have 24 and 53 dimensions, respectively. The focus is on converting the SAC agent's outputs into executable control signals. SAC can output multi-dimensional discrete actions by replacing the Gaussian sampler with a Categorical distribution or Gumbel-SoftMax [67]. However, this approach is not suitable here because temperature setpoints would require a very large number of discrete levels to achieve adequate precision. For example, a single variable would need 51 discrete actions to cover 10 °C to 15 °C at 0.1 °C resolution. Therefore, in this implementation, the agent outputs 24 continuous actions in the range [−1, 1], which are then mapped to the system's physical values. One option is an absolute mapping that linearly scales [−1, 1] to the allowable setpoint range, with intermediate values obtained by linear interpolation. In practice, this mapping has been observed to cause large oscillations within the bounds. For instance, a supply-air temperature of 10 °C at this timestep may jump to 15 °C at the next timestep and then immediately drop to 11 °C. To reflect delays and dead bands in real HVAC equipment, this study uses a relative-incremental mapping. Each action adjusts the previous temperature setpoint by a small, quantized step of 0.1 °C. The agent's output $a \in [-1,1]$ is first linearly scaled and rounded to an increment within [−0.5, 0.5] °C, as shown in Eq. (10). Given the temperature setpoint of the last timestep $T_{t-1}$, the new setpoint for the next timestep $T_t$ can be calculated by Eq. (11). This scheme limits per-step changes, reducing oscillations while preserving sufficient control resolution.

$$\Delta T = 0.1 \times round(5a), \Delta T \in [-0.5, 0.5] \tag{10}$$

$$T_t = T_{t-1} + \Delta T \tag{11}$$

**Reward Function**

The goal of this study is to regulate the aggregated power consumption of a building cluster so that the cluster does not exceed a predetermined power limit threshold. At the same time, the control strategy aims to minimize overall energy consumption while ensuring that the indoor temperatures of all building zones remain within predefined comfort ranges. It is assumed that non-HVAC power demand in the buildings remains constant and is therefore not considered in the analysis. Consequently, the term building power demand in this study refers exclusively to the HVAC power consumption. The reward function is defined in Eq. (12) to minimize the total penalty with three components, power demand penalty $\mathcal{P}_{HVAC}$, thermal comfort penalty $\mathcal{P}_{Temp}(t)$, and peak demand exceedance penalty $\mathcal{P}_{Peak}(t)$. The coefficients are set as $\alpha = 0.5, \beta = 1, \gamma = 2$ through iterative tests based on the control performance.

$$r(t) = -\left( \underbrace{\alpha \mathcal{P}_{HVAC}(t)}_{Power\ Demand} + \underbrace{\beta \mathcal{P}_{Temp}(t)}_{Thermal\ Comfort} + \underbrace{\gamma \mathcal{P}_{Peak}(t)}_{Peak\ Demand} \right) \tag{12}$$

The HVAC power of building $\mathscr{b}$ includes the cooling coil power $P_{\mathscr{b}}^{coil}(t)$ and fan power $P_{\mathscr{b}}^{fan}(t)$ at timestep $t$. Hence, the total power demand of building cluster is shown in Eq. (13). Then, with a given peak demand limiting threshold $P_{max}$, the power demand penalty $\mathcal{P}_{HVAC}(t)$ is calculated by Eq. (14).

$$P_t = \sum_{\mathscr{b}} \left( P_{\mathscr{b}}^{coil}(t) + P_{\mathscr{b}}^{fan}(t) \right) \tag{13}$$

$$\mathcal{P}_{HVAC}(t) = {}^{P_t}/_{P_{max}} \tag{14}$$

The comfort penalty is only considered during occupied hours, i.e., 8 AM to 6 PM. For each controlled zone $z$, an acceptable temperature range $[T_{low}, T_{high}]$ is defined, then the deviation of the current temperature from this range $\Delta T_z(t)$ can be calculated by Eq. (15). To assign stronger penalties to large violations, when the temperature exceeds the range by more than 1 °C (i.e., $\Delta T_z(t) > 1$), $\Delta T_z(t)$ is squared to calculate the comfort penalty for each zone, which is shown in Eq. (16). The total comfort penalty for the building cluster is the sum over all zones, defined as Eq. (17).

$$\Delta T_z(t) = \begin{cases} 0, & T_{low} \leq T_z(t) \leq T_{high} \\ T_{low} - T_z(t), & T_z(t) < T_{low} \\ T_z(t) - T_{high}, & T_z(t) > T_{high} \end{cases} \tag{15}$$

$$\mathcal{P}_z^{temp}(t) = \begin{cases} \Delta T_z(t), & \Delta T_z(t) \leq 1 \\ \left(\Delta T_z(t)\right)^2, & \Delta T_z(t) > 1 \end{cases} \tag{16}$$

$$\mathcal{P}_{Temp}(t) = \sum_z \mathcal{P}_z^{temp}(t) \tag{17}$$

The penalty of peak demand is defined as Eq. (18). If the aggregated power of building cluster is under the given threshold, there would be a positive reward; otherwise, there would be a penalty for violation calculated by the square of the power demand penalty.

$$\mathcal{P}_{Peak}(t) = \begin{cases} (\mathcal{P}_{HVAC}(t))^2, & P_t \geq P_{max} \\ -0.5, & P_t < P_{max} \end{cases} \tag{18}$$

**<u>Hyperparameter Tuning</u>**

To pursue the best possible control performance and the capability of SAC, this study conducted systematic hyperparameter tuning, including: (1) basic hyperparameters were tuned via random search of common values with early stopping that terminated the training after a fixed number of rounds without improvement, yielding ten top-performing combinations; (2) Network architecture was then tuned using an orthogonal array design based on the above ten combinations of basic hyperparameters, running over depth, width, and layer type to evaluate main effects of these parameters within a fixed computational budget; (3) Extended settings that are often kept default in SAC implementation, including weight initialization, activation function and policy distribution, were tuned using grid search. Their effects were discussed on convergence [68], gradient propagation [69], and sample efficiency [70]. This study considered four initializations (Kaiming-Uniform, Kaiming-Normal, Xavier-Uniform, Orthogonal), six activations (ReLU, GELU, ELU, Tanh, Sigmoid, and LeakyReLU with slopes 0.01–0.30), and three policy parameterizations (tanh-squashed Gaussian, unsquashed Gaussian, and Beta) for the tuning. The optimized hyperparameter settings of SAC are summarized in Table 5. It is worth noting that even with identical hyperparameter settings, RL training time and performance cannot be perfectly reproduced due to differences in underlying frameworks and software stacks, hardware (e.g., GPU and drivers), random seeds and other factors [71]. For guidance on reducing nondeterministic behavior, refer to the "Reproducibility" section of the PyTorch documentation [72].

Table 5 Hyperparameter settings of SAC

| | **Hyperparameter** | **Value** |
|---|---|---|
| Basic Hyperparameters | Discount factor | 0.995 |
| | Soft-update coefficient | 0.005 |
| | Target policy entropy | –24 |
| | Actor network learning rate | 0.0003 |
| | Critic network learning rate | 0.0003 |
| | Temperature parameter learning rate | 0.0003 |
| | Optimizer | Adam |
| | Buffer capacity | 10 000 000 |
| | Batch size | 256 |
| Network Architecture | Hidden-layer width | 256 |
| | Hidden-layer count | 2 |
| | Hidden-layer type | Linear (fully-connected) |
| Extended Settings | Normalization | LayerNorm |
| | Weight init | Kaiming-Uniform |
| | Activation | LeakyReLU 0.2 |
| | Policy distribution | Gaussian-tanh |
| | log_std_min (Gaussian) | –20 |
| | log_std_max (Gaussian) | 3 |

# 5. Results and Discussion

## 5.1 Training process of Soft Actor-Critic

Figure 4 presents the evolution of average daily reward across training episodes for the Soft Actor–Critic (SAC) agent compared against the test-day reward. As recorded, the computational running time for each training episode (i.e., 12 days) is about 32 seconds, including RL training and simulations. At the initial phase around 0-500 episodes, the average daily reward per training episode oscillates between –150 and –50, indicating intensive exploration. Throughout this period, the agent receives substantial penalties, resulting from exceeding the aggregated power limit and violating the acceptable range of indoor temperatures. Around 600 episodes, the training curve starts to surpass the test line of baseline (the reward value at –18.78) and continues to ascend, crossing the zero-reward line near 1000 episodes, which suggests that the SAC agent is capable of avoiding the penalties to meet the constraints on power limiting and indoor environment. The training reward thereafter rises steadily, with occasional dips reflecting residual exploration aiming at completely avoiding power-limit breaches. After roughly 2200 episodes, training continues to be improved and narrows the fluctuation till around episode 4500, showing that the SAC algorithm is converged to a stable policy that consistently outperforms the baseline, meeting both comfort and demand-limit objectives, and continues to stabilize the policy. The reward of the test day using the trained SAC agent is 23.45, showing the learned policy is robust to maintain its performance in non-training periods. On the test day, the outdoor temperature peaked at 37.2 °C - higher than the temperature during the training, making control more challenging. But the SAC agent still performs well, demonstrating strong generalization and adaptability to unseen extreme conditions.

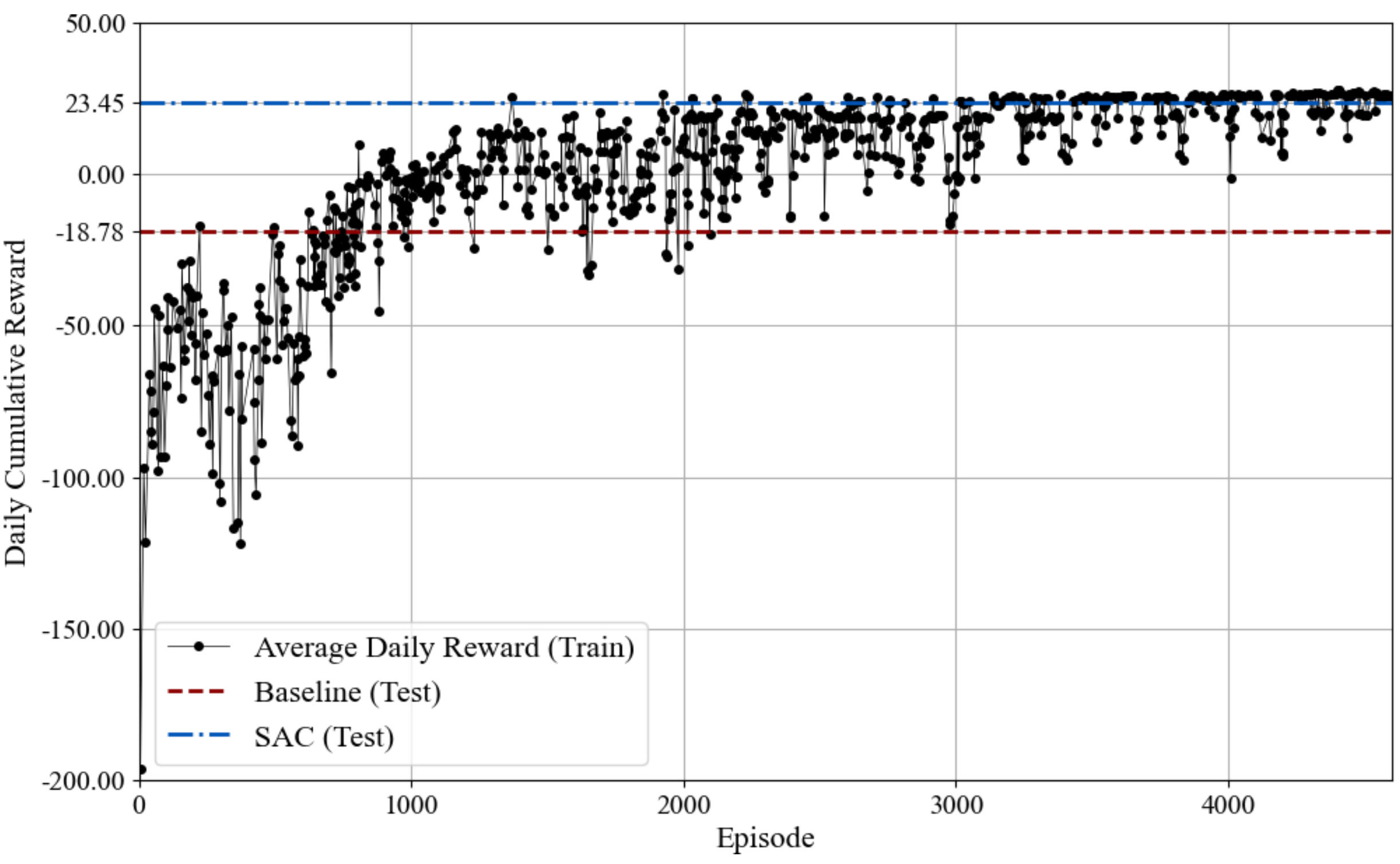

Fig. 4. Training and testing profiles of SAC and baseline

Figure 5 monitors two key training signals for the SAC agent: the adaptive entropy temperature coefficient $\alpha$ and the expected policy log-probability *LogProb*. As defined in Eq. (1), $\alpha$ governs the trade-off between exploration and return. *LogProb* represents the batch-averaged log-probability of the policy for the current action. The log-probability of a single sample is computed by Eq. (7) under the tanh-squashed policy (with the Jacobian term). Its batch-averaged expectation is then used to compute the actor loss in Eq. (8) and adapt $\alpha$ via Eq. (9).

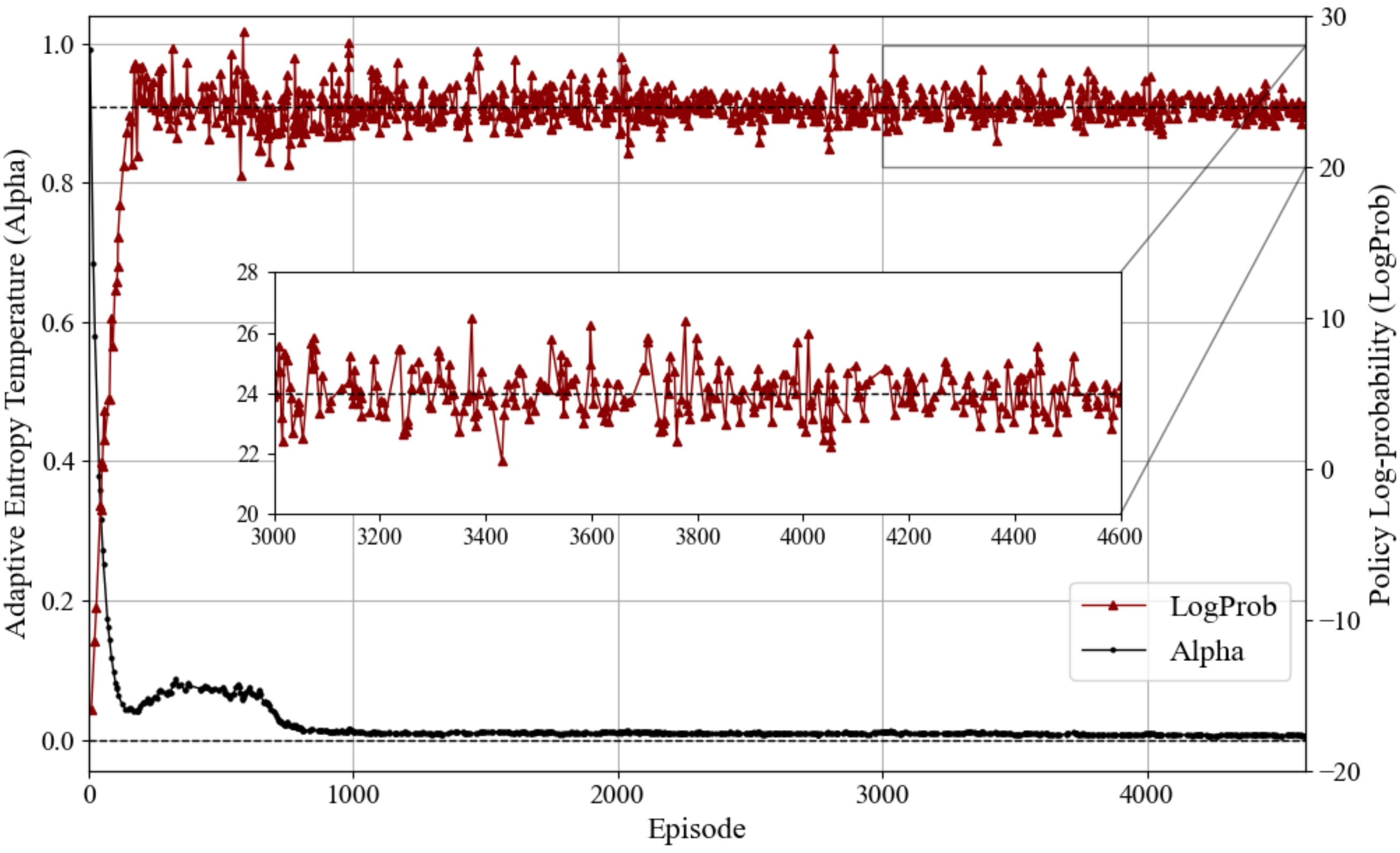

Fig. 5. Adaptive entropy temperature and policy log-probability in SAC training

At the beginning of training, $\alpha$ is initialized at 1 and held at a relatively high level to inject sufficient randomness and encourage exploration. It drops and stabilizes around 0.08 around 600 episodes, which indicates that as the policy matures, the algorithm automatically reduces exploration noise and shifts focus towards exploited behavior. Meanwhile, *LogProb* rapidly climbs from the start of -18 to fluctuate around 24. After 1000 episodes, the fluctuations gradually decrease. This is consistent by design that SAC sets the target entropy as –24, which represents the negative of the 24-dimensional action space (i.e., $\bar{\mathcal{H}} = -|\mathcal{A}|(= -24)$ in Eq. (9)). The adaptation mechanism tunes $\alpha$ by minimizing $\mathcal{L}_\alpha(\alpha)$, which drives the policy's actual entropy approaching the target $\bar{\mathcal{H}}$ (i.e., $\mathbb{E}_{a\sim\pi_\phi} log\pi_\phi(a|s) + \bar{\mathcal{H}} \to 0$). As a result, *LogProb* is observed to naturally stabilize near 24, while the tanh squashing in Eq. (6) and its Jacobian correction in Eq. (7) only add minor corrections without altering the main trend, which can be seen more clearly in the zoomed-in view. The smooth adaptation of $\alpha$ and the convergence of $LogProb$ indicate that exploration noise is being reduced during training with the growing confidence in the current policy, achieving a balanced trade-off between exploration and exploitation.

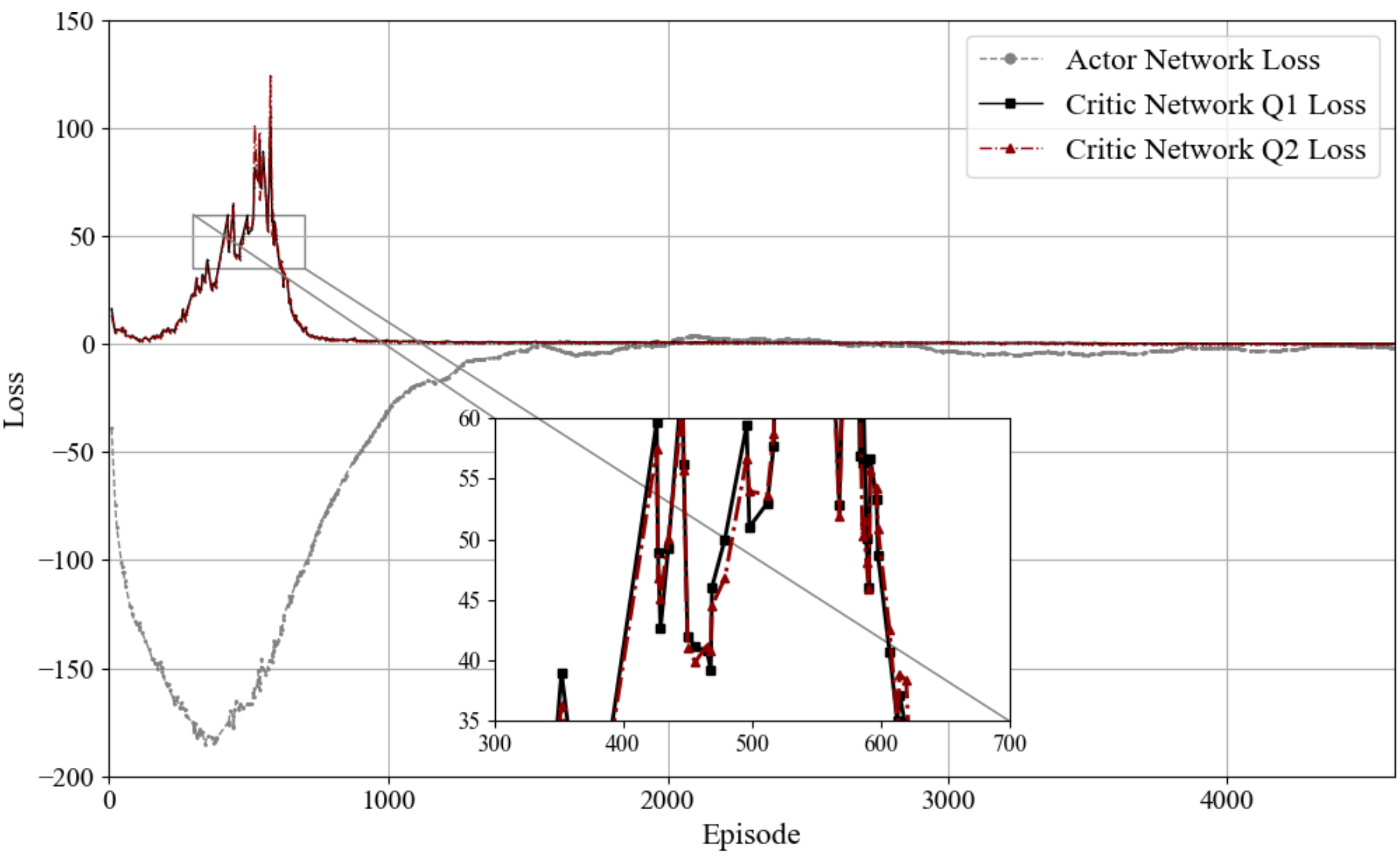

Fig. 6 Actor and double Q-critic losses in the SAC training

Figure 6 tracks the loss curves of the three networks during SAC training. At the first 400 episodes, the actor loss drops to around –180, reflecting large policy updates during exploration; at the same time, the losses of both critic networks rise and peak above 100, indicating substantial bias in estimating environmental returns at the early phase. From roughly 400 to 800 episodes, both critic losses decline and converge toward zero by around 1000 episodes. A zoomed-in view shows that the Q1 and Q2 loss curves have similar trends but do not perfectly overlap. This alternating lead and lag behavior directly illustrates the double Q mechanism at work, since SAC takes the minimum of the two estimates and thereby suppresses overestimation bias. As the critics converge, the actor loss gradually recovers from minimum and stabilizes near zero after 1000 episodes; in later episodes, both actor and critic losses exhibit gentle fluctuations.

## 5.2 Control performance of SAC control agent

As shown in Figure 7, on the test day of July 20th, the aggregated building power demand under baseline control exhibits two peaks: 113.32 kW at around 3 PM and 117.21 kW at around 4 PM. The power limit threshold was set based on the best-performing baseline setting, i.e., the lowest peak demand achieved by the baseline with a 25 °C setpoint. Through experimentation, 103.5 kW was also found to be the minimum level attainable by the SAC algorithm, beyond which further reduction was not achievable. In Figure 7, the SAC agent can keep the aggregated power demand of buildings within the limiting threshold throughout the day, with the peak demand of the day at 4 PM reduced by nearly 12%. Under SAC control, the power demand of the four buildings consistently remains below the baseline during peak hours, demonstrating a coordinated peak-shaving effort with each building contributing varying levels of flexibility. Before 1 PM, SAC control preserves different degrees of pre-cooling based on the indoor temperature conditions of individual zones, resulting in higher overall demand than the baseline for much of this period. Pre-cooling before 8 AM, the office time, stores sufficient cooling resources to release during office hours when necessary to ensure the demand limiting controls.

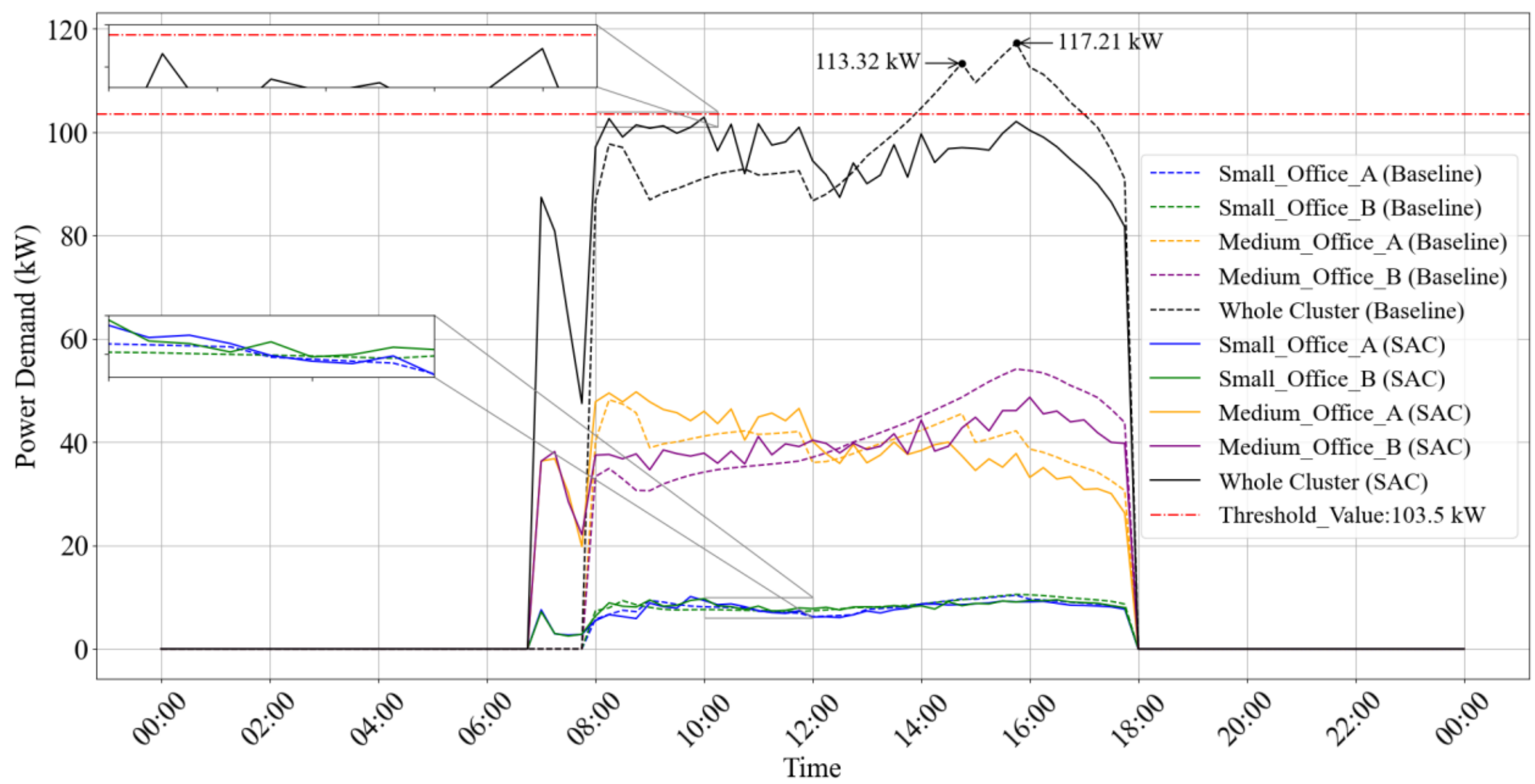

Fig. 7. Power demand profiles of four individual buildings and their aggregation under baseline and SAC controls

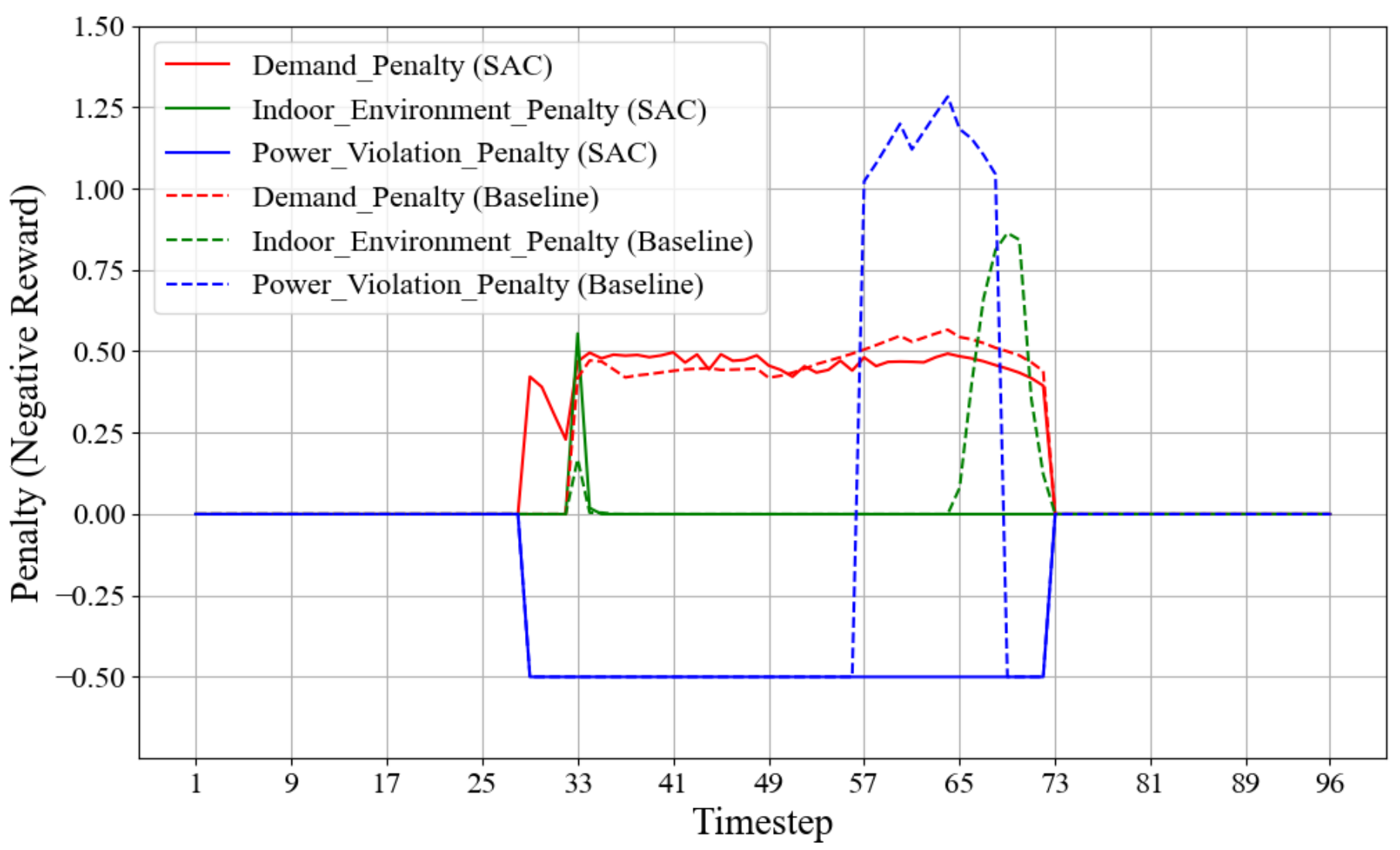

Fig. 8. Breakdown of reward under baseline and SAC controls

Figure 8 shows the breakdown of penalties on the test day of July 20th from power demand, indoor environment and peak demand limiting under baseline and SAC controls. Timestep 1 corresponds to the first interval of 00:15, and the simulation runs for a full day until 24:00 (timestep 96), for a total of 96 timesteps. In Figure 8, the SAC agent consistently achieves the power non-violation reward of 0.5, whereas the baseline controller incurs significant penalties in the afternoon during the threshold exceedance period shown in Figure 7. Using the baseline controller, the comfort violation between timestep 65 (i.e., 16:15) and timestep 72 (i.e., 18:00) is obvious, while the SAC agent is capable of avoiding it. Both controllers exhibit small morning spikes in indoor environment penalties because the bottom floor of the medium office requires less cooling load at this period, resulting in approximately 0.2 °C overcooling (i.e., zone temperature around 22.8 °C). Overall, the baseline controller violates the threshold of power demand limiting between 14:15 and 17:00 and accumulates approximately 43.15 in the indoor environment penalty throughout the day. In contrast, SAC strictly maintains the power demand under the limiting threshold and reduces the total indoor environment penalty to 5.74.

Figure 9 shows the indoor temperature profiles at the zone level of four buildings on the test day of July 20th. During the morning precooling phase from 7 AM to 8 AM, the indoor temperatures are lowered to store cooling energy in the building thermal mass. Although this phase occurs during non-office hours, the lower bound of the temperature setpoint is kept the same as during occupied hours. This ensures thermal comfort in case occupants arrive earlier than the start of office time. After the precooling phase, indoor temperatures gradually rise to minimize energy consumption while ensuring sufficient cooling energy is available for release during peak demand-limiting periods. During the peak period, most of the indoor temperature approaches the upper bound of the comfort range to maximize the release of stored cooling energy, helping to keep the aggregated building demand below the specified threshold.

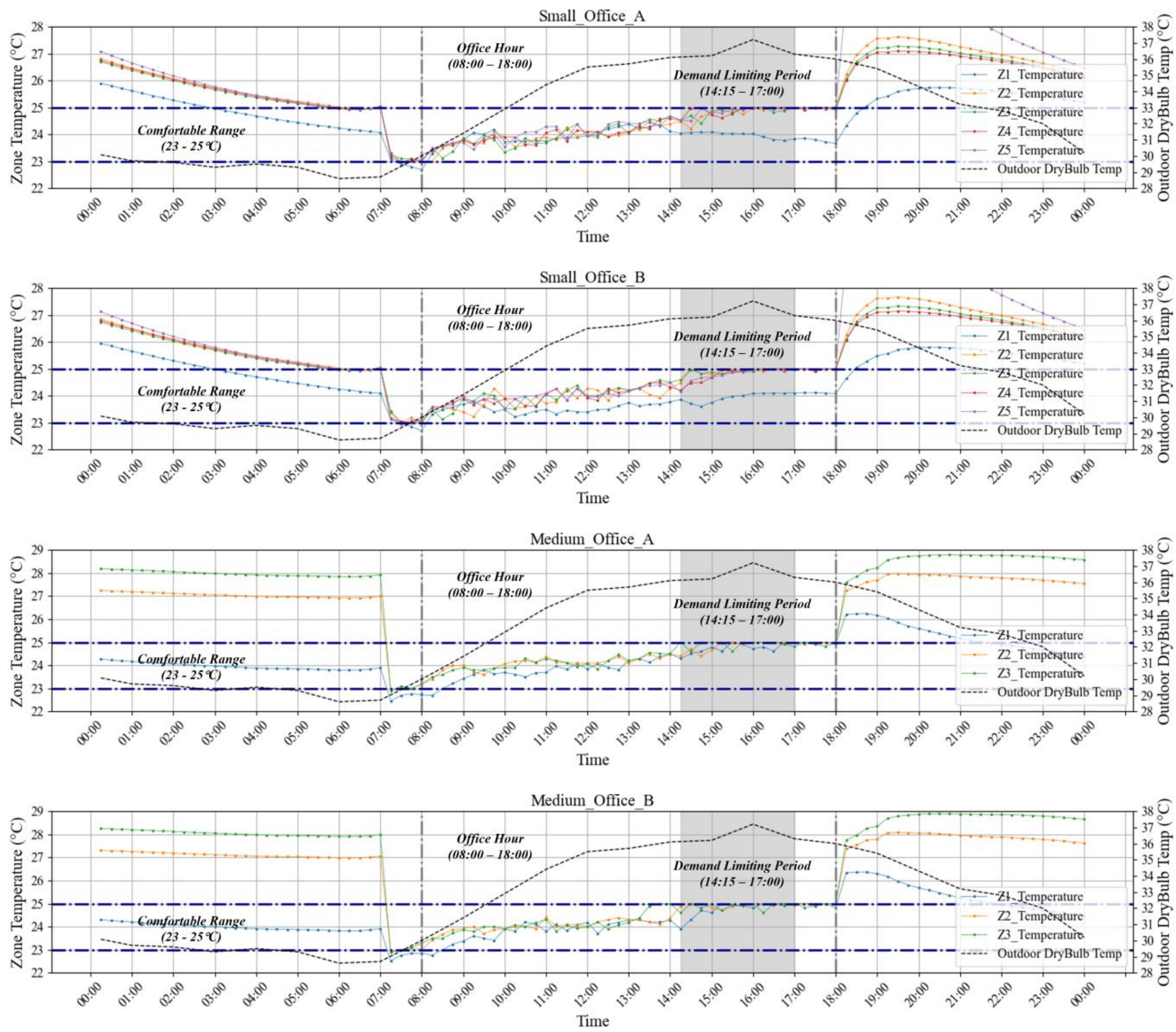

Fig. 9. Reset indoor temperature profiles of each zone/floor in four buildings under SAC controls

All zones exhibit a temperature increase after the cooling system is turned off at 6 PM. The outdoor temperature is approximately 36 °C, decreasing only gradually and remaining at 30.2 °C by midnight. In the small offices, Zone 1 maintains a lower indoor temperature due to its south-facing orientation, lower occupancy, and the use of window shading, as shown in Table 2. In contrast, Zone 5 - the largest zone and located at the center of the building - experiences a more rapid temperature rise after 6 PM. In the medium offices, Zone 1 (located on the bottom floor) receives minimal solar radiation because it is shaded by the building structure, and therefore does not exhibit a significant temperature increase after 6 PM. Overall, the control agent can strategically leverage the building's thermal mass to release stored cooling at a controlled rate to realize the expectation of power demand management.

## 5.3 Coordinated operation of HVAC systems in buildings

Figure 10 presents the supply air flow and fan power in each building on the test day of July 20th. Under SAC control, the supply air flow rates remain stable, with fan power consistently lower than that of the baseline controller across all four buildings. In the small offices, SAC maintains nearly flat air flow rates from system startup through the demand limiting period, whereas the baseline rapidly rises above 800 W. In the medium offices, the supply air flows to all zones show minimal fluctuations under SAC control. During the peak demand period, total fan power is approximately 1000 W and 2000 W lower than the baseline in medium office A and B, respectively. These results indicate that by adjusting the setpoints of indoor temperature and the supply air temperature of air handling unit (AHU), the SAC agent can effectively manage the complex coupling among HVAC components and minimize the fan power by reducing the airflow delivery to achieve energy savings.

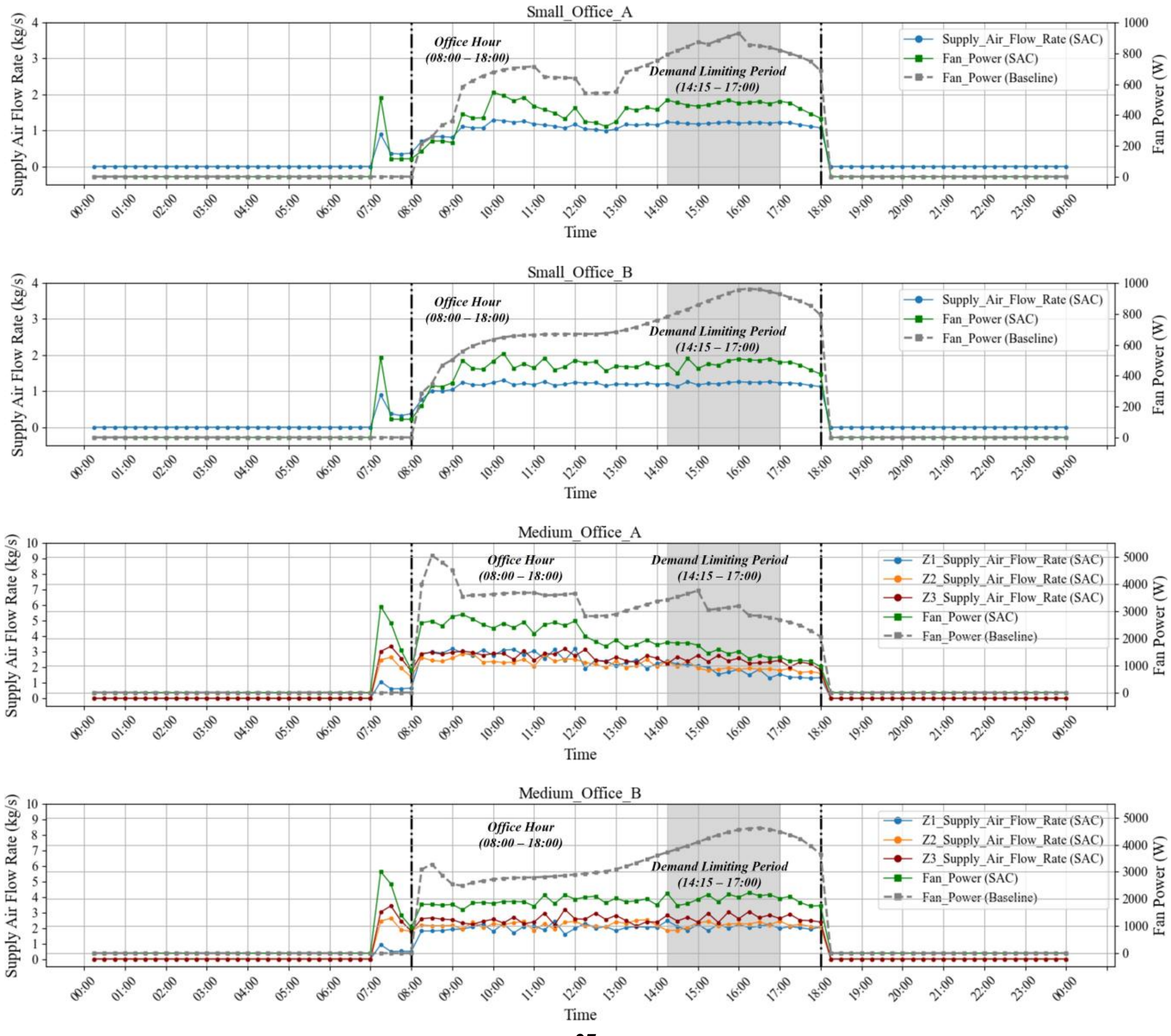

Fig. 10. Supply air flow and fan power of cooling systems under baseline and SAC controls

Figure 11 shows the AHU supply air temperature and cooling coil power of four office buildings on the test day of July 20th. The minimum AHU supply air temperature setpoint is set as 10 °C. Once entering the demand limiting period, the cooling coil power of all four buildings under SAC control remains below the baseline by releasing the cooling stored in buildings. The SAC agent lowers the supply air temperature to reduce the flow rate and fan power to ensure the power demand below the threshold during peak periods. The AHU damper position in Zone 4 of Small Office A, as shown in Figure 12, is further checked. It can be seen that in response to the cooling shortfall caused by high outdoor temperatures in the afternoon, the baseline controller continues increasing the damper opening until it becomes fully open after 4 PM. Because the supply air temperature of baseline controller remains fixed at a high value (i.e., 15 °C), the system cannot meet the cooling load, leading to the indoor environment penalty observed in Figure 8 from timestep 65 (i.e., 16:15) to timestep 72 (i.e., 18:00). In contrast, the SAC controller lowers the supply air temperature in advance, allowing cooling demand to be met with proper damper positions and avoiding unnecessary increases in fan power, thereby preventing indoor environment penalty.

The above analysis highlights the advantages of MuFlex, a physics-based platform. It can enable the zone-level control within a building and provide full access to intermediate variables, which cannot be realized by existing grey-box (e.g., RC) or data-driven models. It can be seen that in this case study under the baseline controller, a particular west-facing room (Zone 4 in Small Office A) experiences indoor temperatures above the comfort range due to high solar radiation in the afternoon, which is very common in real-world scenarios. Zone-level control directly addresses this practical issue by allowing the controller to target thermally critical zones. In addition, the full access to the information of HVAC operations provided by MuFlex, such as supply air temperature of cooling coil, zone airflow delivered by fans, damper position for each zone, supports the development and interpretation of SAC policy. This increases the credibility of control policies trained and provides a more reliable basis for real-world deployments.

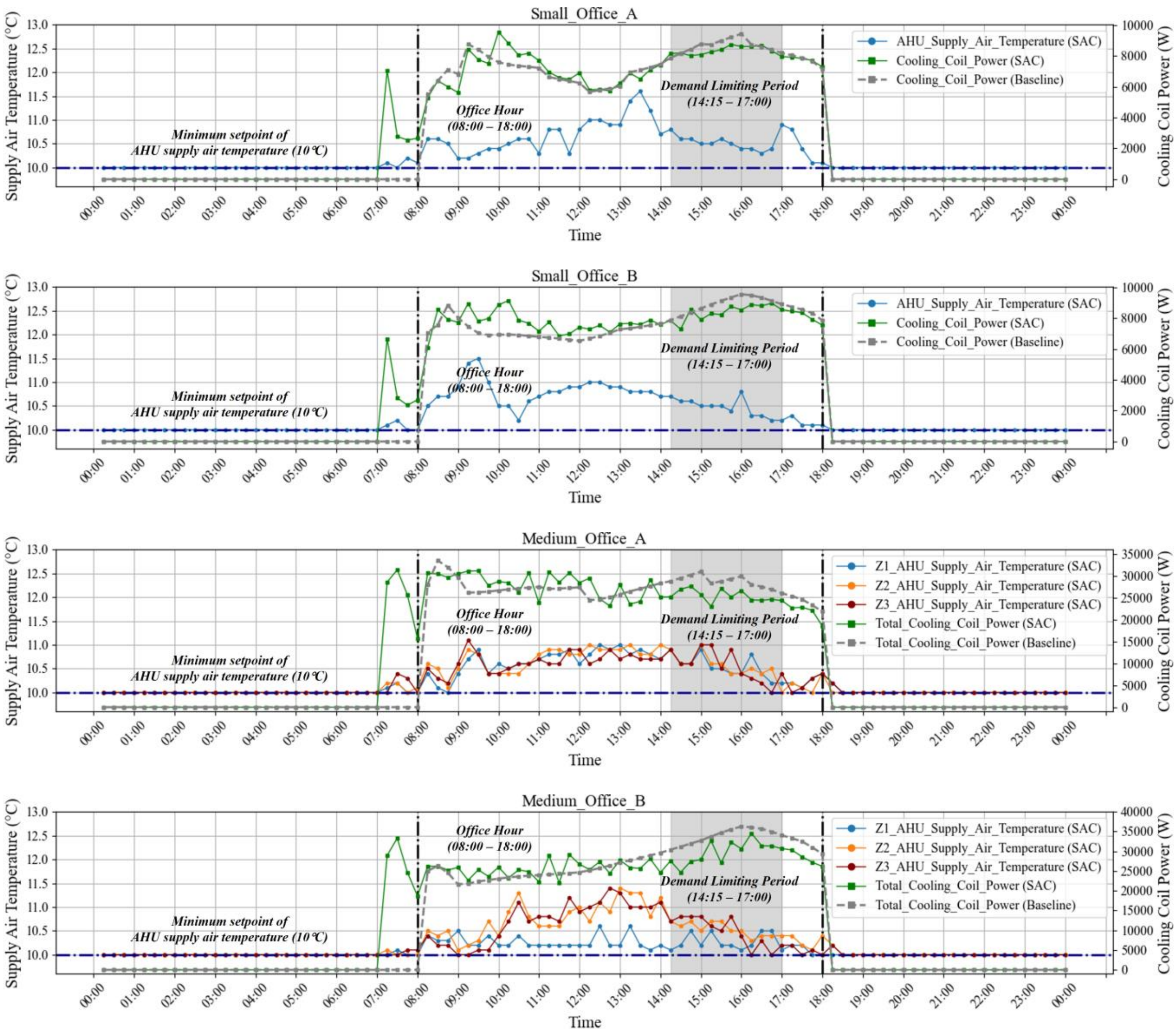

Fig. 11. Supply air temperature and coil power of cooling systems under baseline and SAC

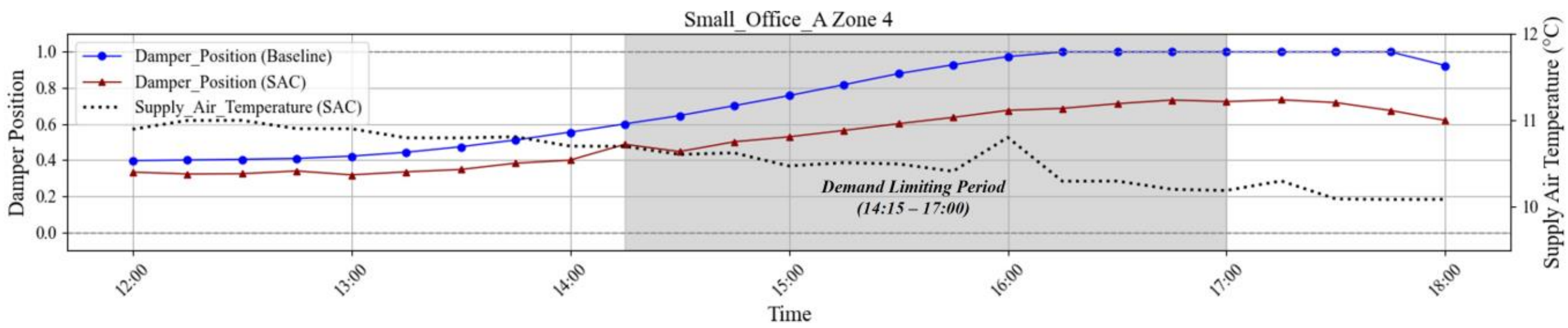

Fig. 12. AHU damper position for Zone 4 in Small Office A

## 5.4 Scalability test of MuFlex

To demonstrate the scalability of MuFlex, a scalability test was conducted across different building clusters. In addition to the office buildings used in the previous case study, a residential house model from the Energym [37], a single-building platform, was incorporated. This residential model was built with the Modelica language and represented a heat pump system. The control input of this model was the normalized heat pump power, and the 22 outputs included 7 weather variables, 12 HVAC operating states, and 3 indoor temperatures. This model was included to highlight that MuFlex could integrate the diverse FMU-compliant building models without being constrained by specific building types or a particular simulation engine.

Table 6 demonstrates the recorded runtime and memory usage for clusters of 4, 20, and 50 buildings simulated over one month (30 days). Case I used the four-building cluster from the case study as the reference. Case II extended Case I by increasing each office model from 1 to 5 (from 4 to 20 buildings in total) to examine how computational cost scaled with cluster size. Case III kept the cluster size at 20 but replaced 10 Medium Office models with 10 Modelica residential models to assess whether mixing simulation engines affected scalability. Case IV included 10 of each model (50 buildings in total) to evaluate scalability in a larger cluster. To ensure fair comparisons across cases, the baseline controller from the case study was used in all four cases to drive the simulation through the standard RL-environment interactions, so that the reported runtime and memory reflected the platform's simulation cost rather than RL cost, which varies with different neural networks and GPU settings.

All tests were executed on the same laptop reported in Section 4.1. Memory Usage (MB) was calculated as the increase in the process RAM usage, measured once before and after each run. This value reflects the memory of the whole process, including the simulation engines (e.g., Modelica and EnergyPlus), and local libraries or compiled extensions (e.g., NumPy, PyFMI, and C/C++ code embedded within FMUs). The peak memory allocated by the Python itself was recorded and reported as Peak Memory (MB).

Table 6 Scalability test of MuFlex: runtime and memory usage across different building clusters

| | Small Office (number) | | Medium Office (number) | | Residential House (number) | Runtime (s) | Memory Usage (MB) | Peak Memory (MB) |
|---|---|---|---|---|---|---|---|---|
| | A | B | A | B | | | | |
| ***Case I.*** 4 Buildings | 1 | 1 | 1 | 1 | 0 | 47 | 303 | 57 |
| ***Case II.*** 20 Buildings | 5 | 5 | 5 | 5 | 0 | 248 | 1303 | 276 |
| ***Case III.*** 20 Buildings, | 5 | 5 | 0 | 0 | 10 | 241 | 1281 | 295 |
| ***Case IV.*** 50 Buildings, | 10 | 10 | 10 | 10 | 10 | 632 | 2319 | 729 |

As can be seen from Table 6, the heaviest simulation of 50 buildings for 30 days (Case IV.) only took 632 seconds and consumed 2319 MB of memory. This was a relatively small computational burden for a laptop with 32 GB of memory. The runtime and memory usage were approximately linear with respect to the number of buildings and simulation days. Within the same 30 days, from Case III. 20 buildings to Case IV. 50 buildings (2.5 times), the increase in runtime was approximately 2.6 times, while the increase in memory usage was only about 1.8 times. Peak memory mainly reflects the accumulation of timestep data stored on the Python side, which also grew linearly and steadily with the number of steps.

By comparing the two 20-building settings (Case II. and Case III.), it was observed that introducing the Modelica-based residential model did not undermine scalability, as the performance remains nearly unchanged. The scalability observed above can be attributed to the platform's ability to leverage the inherent features of simulation programs to run in parallel. In other words, it did not need to wait for one building to finish before advancing another. Consequently, MuFlex supports scalable building clusters with diverse building types and sizes, enables co-simulation across different simulation engines, and facilitates plug-and-play integration of community models with minimal effort.

## 6. Conclusions

To facilitate demand response aggregation across multiple buildings, this study developed MuFlex, an open-source simulation platform designed to benchmark, test, and analyze control optimization strategies for coordinating multi-building flexibility. The platform enables synchronous communication among multiple white-box models (e.g., EnergyPlus and Modelica models), each simulating the physical processes of individual buildings at zone levels and has flexible model interface to configure the inputs and outputs. It captures building thermal dynamics in detail, enabling in-depth analysis of control strategies through the inspection of intermediate variables of buildings. This level of detail also facilitates the trustworthy advanced controls and their transferability to real-world applications. MuFlex integrates with the latest OpenAI Gym interface, providing seamless access to state-of-the-art reinforcement learning libraries. This integration simplifies the implementation of advanced control techniques such as model predictive control and reinforcement learning. A dedicated communication hub was developed for the platform to support real-time interactions between control agents and the physical systems of building clusters. It automatically synchronizes co-simulation across multiple models from diverse simulators using Functional Mock-up Interface (FMI) and leverages parallel model execution to improve computational efficiency of large-scale simulation.

The physical foundation establishes MuFlex as a standardized platform to enable diverse scenario applications for multi-scale research and development in realistic building controls. A case study was conducted using MuFlex to coordinate demand flexibility across four office buildings via the Soft Actor–Critic (SAC) algorithm. The study demonstrated the platform's physics-based capabilities and whole workflow for training, benchmarking, and performing comprehensive control performance analysis. Results showed that a well-tuned SAC agent successfully maintained buildings' efficient operation and thermal comfort while effectively limiting aggregated demand below a defined threshold by leveraging building thermal mass. The scalability of MuFlex was investigated through computational benchmarking on simulating building clusters of different sizes and building types. Results showed modest linear increases in runtime and memory usage as the cluster size grew, while the computational efficiency remained consistent when Modelica-based residential models were included.

Future work will extend MuFlex with more representative metrics and a broader set of baseline algorithms to support systematic benchmarking across diverse control scenarios, including multi-agent RL under different control schemes (e.g., decentralized and distributed) to support coordination across large-scale building clusters. In addition, building models validated or calibrated against real buildings will be provided progressively, and a more rigorous workflow of using MuFlex to build more realistic multi-building control scenarios will be presented.

## Code Availability

The source code is publicly available on GitHub: https://github.com/BuildNexusX/MuFlex.

## Acknowledgements

This work was supported by the University College London Global Engagement Fund and the Engineering and Physical Sciences Research Council (EPSRC) grant (EP/X525649/1).

## References

[1] Wang H, Wang S, Tang R. Development of grid-responsive buildings: Opportunities, challenge s, capabilities and applications of HVAC systems in non-residential buildings in providing anci llary services by fast demand responses to smart grids. Appl Energy 2019;250:697–712. https:// doi.org/10.1016/J.APENERGY.2019.04.159.

[2] Shi J, Zhang G, Liu X. Generation Scheduling Optimization of Wind-Energy Storage Generation System Based on Feature Extraction and MPC. Energy Procedia 2019;158:6672–8. https://doi.org/10.1016/J.EGYPRO.2019.01.028.

[3] Jabir HJ, Teh J, Ishak D, Abunima H. Impacts of Demand-Side Management on Electrical Power Systems: A Review. Energies 2018, Vol 11, Page 1050 2018;11:1050. https://doi.org/10.3390/EN11051050.

[4] Van Heerden R, Edelenbosch OY, Daioglou V, Le Gallic T, Baptista LB, Di Bella A, et al. Demand-side strategies enable rapid and deep cuts in buildings and transport emissions to 2050. Nat Energy 2025;10(3):380–94. https://doi.org/10.1038/s41560-025-01703-1.

[5] Che WW, Tso CY, Sun L, Ip DYK, Lee H, Chao CYH, et al. Energy consumption, indoor thermal comfort and air quality in a commercial office with retrofitted heat, ventilation and air conditioning (HVAC) system. Energy Build 2019;201:202–15. https://doi.org/10.1016/J.ENBUILD.2019.06.029.

[6] Tang R, Wang S. Model predictive control for thermal energy storage and thermal comfort optimization of building demand response in smart grids. Appl Energy 2019;242:873–82. https://doi.org/10.1016/J.APENERGY.2019.03.038.

[7] Li W, Abu bakr M. Assessing the economic feasibility of renewable integration in district heating and cooling networks for sustainable urban development. Energy 2025;339:138814. https://doi.org/10.1016/J.ENERGY.2025.138814.

[8] Hu M, Xiao F, Wang S. Neighborhood-level coordination and negotiation techniques for managing demand-side flexibility in residential microgrids. Renew Sustain Energy Rev 2021;135:110248. https://doi.org/10.1016/J.RSER.2020.110248.

[9] Tang R, Wang S, Li H. Game theory based interactive demand side management responding to dynamic pricing in price-based demand response of smart grids. Appl Energy 2019;250:118–30. https://doi.org/10.1016/J.APENERGY.2019.04.177.

[10] Pinto G, Piscitelli MS, Vázquez-Canteli JR, Nagy Z, Capozzoli A. Coordinated energy management for a cluster of buildings through deep reinforcement learning. Energy 2021;229. https://doi.org/10.1016/j.energy.2021.120725.

[11] Jensen SØ, Marszal-Pomianowska A, Lollini R, Pasut W, Knotzer A, Engelmann P, et al. IEA EBC Annex 67 Energy Flexible Buildings. Energy Build 2017;155:25–34. https://doi.org/10.1016/J.ENBUILD.2017.08.044.

[12] Documents | ASHRAE 36 High Performance Sequences of Operation for HVAC Systems n.d. https://tpc.ashrae.org/Documents?cmtKey=d536fedd-5057-4fc6-be3a-808233902f4c (accessed May 28, 2025).

[13] Wang Z, Hong T. Reinforcement learning for building controls: The opportunities and challenges. Appl Energy 2020;269:115036. https://doi.org/10.1016/J.APENERGY.2020.115036.

[14] Tang R, Fan C, Zeng F, Feng W. Data-driven model predictive control for power demand management and fast demand response of commercial buildings using support vector regression. Build Simul 2022;15:317–31. https://doi.org/10.1007/S12273-021-0811-X.

[15] Mnih V, Kavukcuoglu K, Silver D, Rusu AA, Veness J, Bellemare MG, et al. Human-level control through deep reinforcement learning. Nature 2015;518:529–33. https://doi.org/10.1038/NATURE14236.

[16] Vinyals O, Babuschkin I, Czarnecki WM, Mathieu M, Dudzik A, Chung J, et al. Grandmaster level in StarCraft II using multi-agent reinforcement learning. 350 | Nat | 2019;575. https://doi.org/10.1038/s41586-019-1724-z.

[17] Release Notes - Gymnasium-Robotics Documentation n.d. https://robotics.farama.org/release_notes/ (accessed May 28, 2025).

[18] Haider M, Yin M, Zhang M, Gupta A, Barbara S, Zhu J, et al. NetworkGym: Reinforcement Learning Environments for Multi-Access Traffic Management in Network Simulation. Adv Neural Inf Process Syst 2024;37:106628–48.

[19] Gao Y, Shi S, Miyata S, Akashi Y. Successful application of predictive information in deep reinforcement learning control: A case study based on an office building HVAC system. Energy 2024;291:130344. https://doi.org/10.1016/j.energy.2024.130344.

[20] Gymnasium Documentation n.d. https://gymnasium.farama.org/ (accessed June 11, 2025).

[21] Raffin A, Hill A, Gleave A, Kanervisto A, Ernestus M, Dormann N. Stable-Baselines3: Reliable Reinforcement Learning Implementations. J Mach Learn Res 2021;22:1–8.

[22] Nagy Z, Henze G, Dey S, Arroyo J, Helsen L, Zhang X, et al. Ten questions concerning reinforcement learning for building energy management. Build Environ 2023;241:110435. https://doi.org/10.1016/J.BUILDENV.2023.110435.

[23] Lyu Z, Peng X, Zhang Y, Zhang B, Lin B, Yuan C, et al. Knowledge-guided reinforcement learning for HVAC controls and energy saving through an efficient simulation framework. Energy 2025;339:139009. https://doi.org/10.1016/J.ENERGY.2025.139009.

[24] Cui C, Xue J. Energy and comfort aware operation of multi-zone HVAC system through preference-inspired deep reinforcement learning. Energy 2024;292:130505. https://doi.org/10.1016/J.ENERGY.2024.130505.

[25] Zhang Z, Lam KP. Practical implementation and evaluation of deep reinforcement learning control for a radiant heating system. BuildSys 2018 - Proc 5th Conf Syst Built Environ 2018:148–57. https://doi.org/10.1145/3276774.3276775.

[26] Lu X, Fu Y, O'Neill Z. Benchmarking high performance HVAC Rule-Based controls with advanced intelligent Controllers: A case study in a Multi-Zone system in Modelica. Energy Build 2023;284:112854. https://doi.org/10.1016/J.ENBUILD.2023.112854.

[27] Yuan X, Pan Y, Yang J, Wang W, Huang Z. Study on the application of reinforcement learning in the operation optimization of HVAC system. Build Simul 2021;14:75–87. https://doi.org/10.1007/S12273-020-0602-9.

[28] EMS Application Guide — EnergyPlus 23.1 documentation n.d. https://energyplus.readthedocs.io/en/latest/ems-application-guide/ems-application-guide.html (accessed May 28, 2025).

[29] EnergyPlus Python API — EnergyPlus 23.1 documentation n.d. https://energyplus.readthedocs.io/en/latest/api.html (accessed May 28, 2025).

[30] lbl-srg/bcvtb: Building Controls Virtual Test Bed n.d. https://github.com/lbl-srg/bcvtb (accessed May 28, 2025).

[31] Functional Mock-up Interface n.d. https://fmi-standard.org/ (accessed May 28, 2025).

[32] Zhang T, Ardakanian O. COBS: COmprehensive Building Simulator. BuildSys 2020 - Proc 7th ACM Int Conf Syst Energy-Efficient Build Cities, Transp 2020:314–5. https://doi.org/10.1145/3408308.3431119.

[33] Moriyama T, De Magistris G, Tatsubori M, Pham TH, Munawar A, Tachibana R. Reinforcement learning testbed for power-consumption optimization. Methods and Applications for Modelin

g and Simulation of Complex Systems, AsiaSim 2018 Proceedings. Communications in Comput er and Information Science, vol. 946. Springer; 2018. p. 45–59. https://doi.org/10.1007/978-981-13-2853-4_4.

[34] airboxlab/rllib-energyplus: Simple EnergyPlus environments for control optimization using reinfo rcement learning n.d. https://github.com/airboxlab/rllib-energyplus/tree/main (accessed May 8, 20 25).

[35] Campoy-Nieves A, Manjavacas A, Jiménez-Raboso J, Molina-Solana M, Gómez-Romero J. Sin ergym – A virtual testbed for building energy optimization with Reinforcement Learning. Ener gy Build 2025;327:115075. https://doi.org/10.1016/J.ENBUILD.2024.115075.

[36] Zhang Z, Chong A, Pan Y, Zhang C, Lam KP. Whole building energy model for HVAC opti mal control: A practical framework based on deep reinforcement learning. Energy Build 2019;1 99:472–90.

[37] Scharnhorst P, Schubnel B, Fernández Bandera C, Salom J, Taddeo P, Boegli M, et al. Energ ym: A Building Model Library for Controller Benchmarking. Appl Sci 2021, Vol 11, Page 35 18 2021;11:3518. https://doi.org/10.3390/APP11083518.

[38] WalterZWang/AlphaBuilding-MedOffice: This is the official repository of AlphaBuilding MedOf fice: a realistic OpenAI Gym environment that can be used to train, test and benchmark contr ollers for medium size office (1AHU + 9VAV boxes) n.d. https://github.com/WalterZWang/Alp haBuilding-MedOffice (accessed May 28, 2025).

[39] Touzani S, Prakash AK, Wang Z, Agarwal S, Pritoni M, Kiran M, et al. Controlling distribute d energy resources via deep reinforcement learning for load flexibility and energy efficiency. Appl Energy 2021;304:117733. https://doi.org/10.1016/J.APENERGY.2021.117733.

[40] Marzullo T, Dey S, Long N, Leiva Vilaplana J, Henze G. A high-fidelity building performanc e simulation test bed for the development and evaluation of advanced controls. J Build Perfor m Simul 2022;15:379–97. https://doi.org/10.1080/19401493.2022.2058091.

[41] Blum D, Arroyo J, Huang S, Drgoňa J, Jorissen F, Walnum HT, et al. Building optimization testing framework (BOPTEST) for simulation-based benchmarking of control strategies in buildi ngs. J Build Perform Simul 2021;14:586–610. https://doi.org/10.1080/19401493.2021.1986574.

[42] Gao C, Wang D. Comparative study of model-based and model-free reinforcement learning con trol performance in HVAC systems. J Build Eng 2023;74:106852.

[43] Wu Z, Zhang W, Tang R, Wang H, Korolija I. Reinforcement learning in building controls: A comparative study of algorithms considering model availability and policy representation. J Bu ild Eng 2024:109497.

[44] Wang D, Zheng W, Wang Z, Wang Y, Pang X, Wang W. Comparison of reinforcement learni ng and model predictive control for building energy system optimization. Appl Therm Eng 202 3;228:120430. https://doi.org/10.1016/J.APPLTHERMALENG.2023.120430.

[45] Tang R, Li H, Wang S. A game theory-based decentralized control strategy for power demand management of building cluster using thermal mass and energy storage. Appl Energy 2019;242: 809–20.

[46] Shen L, Li Z, Sun Y. Performance evaluation of conventional demand response at building-gro up-level under different electricity pricings. Energy Build 2016;128:143–54. https://doi.org/10.10 16/J.ENBUILD.2016.06.082.

[47] Korolija I, Marjanovic-Halburd L, Zhang Y, Hanby VI. Influence of building parameters and HVAC systems coupling on building energy performance. Energy Build 2011;43:1247–53. https://doi.org/10.1016/J.ENBUILD.2011.01.003.

[48] Huang P, Fan C, Zhang X, Wang J. A hierarchical coordinated demand response control for buildings with improved performances at building group. Appl Energy 2019;242:684–94. https://doi.org/10.1016/J.APENERGY.2019.03.148.

[49] Vazquez-Canteli JR, Henze G, Nagy Z. MARLISA: Multi-Agent Reinforcement Learning with Iterative Sequential Action Selection for Load Shaping of Grid-Interactive Connected Buildings. BuildSys 2020 - Proc 7th ACM Int Conf Syst Energy-Efficient Build Cities, Transp 2020:170–9. https://doi.org/10.1145/3408308.3427604.

[50] Vázquez-Canteli JR, Kämpf J, Henze G, Nagy Z. CityLearn v1.0: An OpenAI gym environment for demand response with deep reinforcement learning. BuildSys 2019 - Proc 6th ACM Int Conf Syst Energy-Efficient Build Cities, Transp 2019:356–7. https://doi.org/10.1145/3360322.3360998.

[51] Pigott A, Crozier C, Baker K, Nagy Z. GridLearn: Multiagent reinforcement learning for grid-aware building energy management. Electr Power Syst Res 2022; 213. https://doi.org/10.1016/j.epsr.2022.108521.

[52] Zhang C, Kuppannagari SR, Kannan R, Prasanna VK. Building HVAC scheduling using reinforcement learning via neural network based model approximation. BuildSys 2019 - Proc 6th ACM Int Conf Syst Energy-Efficient Build Cities, Transp 2019:287–96. https://doi.org/10.1145/3360322.3360861.

[53] Mocanu E, Mocanu DC, Nguyen PH, Liotta A, Webber ME, Gibescu M, et al. On-Line Building Energy Optimization Using Deep Reinforcement Learning. IEEE Trans Smart Grid 2019;10:3698–708. https://doi.org/10.1109/TSG.2018.2834219.

[54] Zhang B, Hu W, Ghias AMYM, Xu X, Chen Z. Multi-agent deep reinforcement learning-based coordination control for grid-aware multi-buildings. Appl Energy 2022;328:120215. https://doi.org/10.1016/J.APENERGY.2022.120215.

[55] Pinto G, Deltetto D, Capozzoli A. Data-driven district energy management with surrogate models and deep reinforcement learning. Appl Energy 2021;304:117642. https://doi.org/10.1016/J.APENERGY.2021.117642.

[56] Fonseca T, Ferreira LL, Cabral B, Severino R, Nweye K, Ghose D, et al. EVLearn: extending the cityLearn framework with electric vehicle simulation. Energy Informatics 2025;8:16.

[57] Gallo A, Capozzoli A. The role of advanced energy management strategies to operate flexibility sources in Renewable Energy Communities. Energy Build 2024;325:115043. https://doi.org/10.1016/J.ENBUILD.2024.115043.

[58] Wang Z, Chen B, Li H, Hong T. AlphaBuilding ResCommunity: A multi-agent virtual testbed for community-level load coordination. Adv Appl Energy 2021;4:100061. https://doi.org/10.1016/J.ADAPEN.2021.100061.

[59] Junghanns A, Blochwitz T, Bertsch C, Sommer T, Wernersson K, Pillekeit A, et al. The Functional Mock-up Interface 3.0 - New Features Enabling New Applications. Model Conf 2021;181:17–26. https://doi.org/10.3384/ECP2118117.

[60] Andersson C, Åkesson J, Führer C. PyFMI: A Python package for simulation of coupled dynamic models with the Functional Mock-up Interface. Technical Report 2, LUTFNA-5008-2016, Centre for Mathematical Sciences, Lund University, Lund, Sweden; 2016.

[61] Brockman G, Cheung V, Pettersson L, Schneider J, Schulman J, Tang J, et al. OpenAI Gym. https://doi.org/10.48550/arXiv.1606.01540; 2016.

[62] Liang E, Wu Z, Luo M, Mika S, Gonzalez JE, Stoica I. RLlib Flow: Distributed Reinforcement Learning is a Dataflow Problem. Adv Neural Inf Process Syst 2020;7:5506–17.

[63] EnergyPlusToFMU | Simulation Research n.d. https://simulationresearch.lbl.gov/projects/energyplustofmu (accessed May 28, 2025).

[64] Duan Y, Chen X, Houthooft R, Schulman J, Abbeel P. Benchmarking Deep Reinforcement Learning for Continuous Control. 33rd Int Conf Mach Learn ICML 2016 2016;3:2001–14.

[65] Van Hasselt H, Guez A, Silver D. Deep Reinforcement Learning with Double Q-Learning. Proc AAAI Conf Artif Intell 2016;30:2094–100. https://doi.org/10.1609/AAAI.V30I1.10295.

[66] Haarnoja T, Zhou A, Hartikainen K, Tucker G, Ha S, Tan J, et al. Soft Actor-Critic Algorithms and Applications. https://doi.org/10.48550/arXiv.1812.05905; 2018.

[67] Christodoulou P. Soft Actor-Critic for Discrete Action Settings. https://doi.org/10.48550/arXiv.1910.07207; 2019.

[68] Liu L, Jiang H, He P, Chen W, Liu X, Gao J, et al. On the Variance of the Adaptive Learning Rate and Beyond. 8th Int Conf Learn Represent ICLR 2020 2019.

[69] Lu Z, Pu H, Wang F, Hu Z, Wang L. The Expressive Power of Neural Networks: A View from the Width. Adv Neural Inf Process Syst 2017;2017-Decem:6232–40.

[70] Chou P-W, Maturana D, Scherer S. Improving stochastic policy gradients in continuous control with deep reinforcement learning using the Beta distribution. Proceedings of the 34th International Conference on Machine Learning (ICML 2017); 2017. p. 834–843.

[71] Nagarajan P, Warnell G, Stone P. Deterministic Implementations for Reproducibility in Deep Reinforcement Learning. https://doi.org/10.48550/arXiv.1809.05676; 2018.

[72] Reproducibility — PyTorch 2.7 documentation n.d. https://docs.pytorch.org/docs/stable/notes/randomness.html (accessed May 28, 2025).